\documentclass[times,twocolumn,final,longtitle]{elsarticle}

\usepackage{./medima}
\usepackage{framed,multirow}
\usepackage[utf8]{inputenc}
\usepackage[T1]{fontenc}
\usepackage{graphicx}
\usepackage{amsmath,amssymb,amsfonts}
\usepackage{amsmath}
\newif\ifnotshortening
\notshorteningfalse
\usepackage{float}
\usepackage{textcomp}
\usepackage[noend]{algpseudocode}
\usepackage{booktabs}
\usepackage[ruled,vlined]{algorithm2e}
\usepackage{diagbox}
\usepackage{bigstrut}
\usepackage{dblfloatfix}
\usepackage[graphicx]{realboxes}
\usepackage[table, svgnames, dvipsnames]{xcolor}
\usepackage{stackengine}

\definecolor{blue}{rgb}{0.0, 0.0, 0.0}
\definecolor{green}{rgb}{0.0, 0.0, 0.0}
\definecolor{blue2}{rgb}{0.0, 0.0, 1.0}
\definecolor{revision_1}{rgb}{0,0.0,0.0}
\definecolor{eaai_rev_1}{rgb}{0.0, 0.0, 0.0}
\usepackage{amssymb}
\usepackage{latexsym}
\usepackage{natbib}

\usepackage{url}
\usepackage{xcolor}

\usepackage{hyperref}

\definecolor{newcolor}{rgb}{.8,.349,.1}

\journal{This is the final proofread version submitted to Elsevier EAAI: please see  the published version at: https://doi.org/10.1016/j.engappai.2022.105095}

\begin{document}

\verso{Bilel Benjdira \textit{et~al.}}

\begin{frontmatter}

\title{\textcolor{eaai_rev_1}{TAU: A Framework for Video-Based Traffic Analytics Leveraging Artificial Intelligence and Unmanned Aerial Systems}}%

\author[1]{Bilel  Benjdira}
\ead{bbenjdira@psu.edu.sa}

\author[1]{Anis Koubaa}
\ead{akoubaa@psu.edu.sa}
\author[4,5]{Ahmad Taher  Azar\corref{cor1}}
\ead{aazar@psu.edu.sa}
\cortext[cor1]{Corresponding author: Ahmed Taher Azar (aazar@psu.edu.sa)
  }
\author[1]{Zahid   Khan}
\ead{zskhan@psu.edu.sa}
\author[1]{Adel  Ammar}
\ead{aammar@psu.edu.sa}
\author[1]{Wadii Boulila}
\ead{wboulila@psu.edu.sa}

\address[1]{Robotics and Internet-of-Things Unit (RIOTU) Lab, Prince Sultan University, Riyadh, Saudi Arabia.}
\address[4]{College of Computer and Information Sciences, Prince Sultan University, Riyadh, Saudi Arabia.}
\address[5]{Faculty of Computers and Artificial Intelligence, Benha University, Benha, Egypt.}

\received{""}
\finalform{""}
\accepted{""}
\availableonline{""}
\communicated{" "}

\begin{abstract}
\textcolor{eaai_rev_1}{Smart traffic engineering and intelligent transportation services are in increasing demand from governmental authorities to optimize traffic performance and thus reduce energy costs, increase the drivers' safety and comfort, ensure traffic laws enforcement, and detect traffic violations. In this paper, we address this challenge, and we leverage the use of Artificial Intelligence (AI) and Unmanned Aerial Vehicles (UAVs) to develop an AI-integrated video analytics framework, called TAU (Traffic Analysis from UAVs), for automated traffic analytics and understanding. Unlike previous works on traffic video analytics, we propose an automated object detection and tracking pipeline from video processing to advanced traffic understanding using high-resolution UAV images. TAU combines six main contributions. First, it proposes a pre-processing algorithm to adapt the high-resolution UAV image as input to the object detector without lowering the resolution. This ensures an excellent detection accuracy from high-quality features, particularly the small size of detected objects from UAV images. Second, it introduces an algorithm for recalibrating the vehicle coordinates to ensure that vehicles are uniquely identified and tracked across the multiple crops of the same frame. Third, it presents a speed calculation algorithm based on accumulating information from successive frames. Fourth, TAU counts the number of vehicles per traffic zone based on the Ray Tracing algorithm. Fifth, TAU has a fully independent algorithm for crossroad arbitration based on the data gathered from the different zones surrounding it. Sixth, TAU introduces a set of algorithms for extracting twenty-four types of insights from the raw data collected. These insights facilitate the traffic understanding using curves, histograms, heatmaps, and animations. The present work presents a valuable added value for academic researchers and transportation engineers to automate the traffic video analytics process and extract useful insights to optimize traffic performance.}
\textcolor{eaai_rev_1}{TAU is a ready-to-use framework for any Transportation Engineer to better understand and manage daily road traffic (video demonstrations are provided in these links: \url{https://youtu.be/wXJV0H7LviU} and here: \url{https://youtu.be/kGv0gmtVEbI}.). The source code is shared at this link: \url{https://github.com/bilel-bj/TAU}. }
\end{abstract}

\end{frontmatter}



\section{INTRODUCTION}
In urban areas, people face daily frequent traffic congestion problems. On average, they waste more than 75\% of their usual travel time due to congestion  \citep{olak2016UnderstandingCT}. The situation will be more and more complicated by the increased number of people moving to live in cities. The UN world Urbanization prospects estimated in its 2018 report \citep{UN} that around 2.5 billion people will be living in urban areas by 2050. Hence, the traffic congestion problems are going to be increasingly aggravated. To analyze and monitor the traffic jam, surveillance cameras are placed in the critical parts of the city roads \citep{Haan2010SpatialNF}. Although being informative, these cameras suffer from limited and incomplete vision areas. They can not cover the whole portion of the road in congestion. Moreover, the high cost of installing and maintaining these cameras is also a considerable limitation. Thus, Unmanned Aerial Vehicles (UAVs) are an attractive tool for analyzing traffic. This is why the authorities are increasingly adopting them all over the world. This increasing adoption is due to many factors. First, they are flexible and could be rapidly flown over the congestive portion of the roads to take a real-time video of the traffic area. Second, they need no additional installation or maintenance cost for this analysis mission. Third, they could easily be connected to 5G networks to stream the captured videos in high bandwidth while maintaining high resolution. Hence, the UAVs possess a high potential to be used for various tasks related to traffic management and analysis. \par
\textcolor{revision_1}{ The advantages of using UAVs does not imply giving up the use of stationary cameras. There are many scenarios in which stationary cameras are more appropriate than UAVs. Indeed, UAVs suffer from battery constraints that hamper their adoption for continual use. Also, Ensuring stable temporal coverage by using only UAVs is challenging. Moreover, bad weather conditions may complicate their operation. Besides, the safety and the compliance with different civil aviation laws is another limitation to face. Hence, it is better to combine stationary cameras and UAVs to overcome the limitations of both tools. This improves the robustness of traffic analysis and management. However, before studying the possibility of this combination, this paper aims to profoundly study the potential of using only UAVs in this task. } \par
Currently, most use cases for the UAVs inside the roads are deprived of machine intelligence. The majority of the efforts are focused on gathering real-time videos of the traffic for visual analysis and investigation by an engineer. The engineer will train himself to recognize the causes of traffic violations or how to choose suitable measures to reduce congestion. He will decide on more challenging problems like how to optimize the traffic signals and the time slots allocated for every road portion. More critical situations can face the engineer, such as the accident risk assessment. He will have to report to the auto insurance industry and the police departments to help them judge the potential responsibility of the counterparties. These are some of the challenging missions that create an insistent need for an automatic tool that alleviates the process and makes the monitoring and the management more efficient and less prone to errors. \par
On the other hand, Since 2012 \citep{alexnet}, an impressive advancement in computer vision applications has been noted. This advancement is due to the appearance of the area of deep learning algorithms that made image and video processing more efficient than ever before \citep{goodfellow2016deep,liu2020deep,Benjdira2019Segmentation, data-efficient,spinal-cord,Alhichri_2019, alhichri2016multiple, koubaa2020activity,boulila2021deep, alkhelaiwi2021efficient, ben2022randomly,benjdira2022parking,koubaa2020deepbrain,noor2020driftnet,ben2021covid}. Many deep learning algorithms have been incrementally optimized for the main tasks on deep learning such as object detection inside images \citep{liu2020deep}, object tracking in videos \citep{ciaparrone2020deep}, instance segmentation \citep{hafiz2020survey} and semantic segmentation \citep{hao2020brief}. \par
\textcolor{eaai_rev_1}{These advancements in deep learning have inspired a variety of research projects that aimed at maximizing the use of UAV images to improve traffic services \citep{OUTAY2020116}.}
\textcolor{eaai_rev_1}{However, the variety of studies and techniques hardens the mission for a transportation engineer to pick out the best technique for every insight he wants to extract from the video. Moreover, he will face difficulties in testing the feasibility of every technique apart. Besides, integrating the selected techniques will be another issue to resolve.} \par
\textcolor{eaai_rev_1}{This paper aims to target this matter and to design a unified and straightforward pipeline for the extraction of a comprehensive set of insights from the UAV video (more than 20 types of insights). 
The founding Framework for this pipeline is coined TAU (Traffic Analysis from UAVs). It helps the Transportation Engineer to automatically analyze the traffic from a UAV video in a seamless way. To summarize, TAU introduces six main contributions:
\begin{itemize}
\item The first contribution is a new pre-processing algorithm to enable the object detector to work in parallel on high-resolution images. The algorithm decomposes the input frame into multiple small crops to keep the ability to detect the small-sized vehicles. 
\item The second contribution is a new algorithm to recalibrate the vehicle coordinates so that the multi-object tracker can efficiently identify all the circulating vehicles in the high-resolution video. 
\item The third contribution made is a new algorithm to calculate every vehicle's speed in the video. The mathematical formulation behind this algorithm is explicitly detailed. 
\item The fourth contribution is a new algorithm to calculate the number of vehicles per traffic zones. The algorithm follows the Ray Tracing approach in treating this problem. 
\item The fifth contribution is a new crossroad arbitration algorithm. This algorithm is based on the data gathered from the different traffic zones surrounding it. 
\item And finally, the sixth contribution is a set of algorithms for extracting twenty-four types of insights from the traffic video. These insights enable better traffic understanding and analysis using histograms, heatmaps, curves, and animations. The TAU framework was a successful research prototype designed in an industrial context and opened later to the community.
\end{itemize}}\par 
This paper is structured as follows: Section \ref{section:related-work} will describe the research works related to the use of UAVs inside the domain of traffic analysis and management. Then, section \ref{section:proposed-method} is reserved for the theoretical founding of the TAU framework. Later, Section \ref{section:experimental-results} describes the implementation details made to test TAU as well as the generated results. Finally, section \ref{section:conclusion} concludes the utility of TAU and describes the limitations to address in future works. 

\section{Related Works}\label{section:related-work}
This section inspects the research works that used UAV imagery to enhance traffic monitoring and management. The majority of these research works are focused on three classes \citep{ke2018real}. The first class is road detection; the second class is vehicle detection and tracking; the third class is traffic parameter estimation. A description of the research works implemented in every class will be detailed in the following subsections.
\subsection{Road detection}

Road detection from UAV was the target of many studies in literature \citep{ghandorh2022semantic, zhou2014efficient, lin2012road, kim2005realtime, pless2004road}. The aim was to localize the areas where the different types of vehicles circulate. This goal is essential to automatically make the UAV focus on the areas of interest and neglect other regions. \textcolor{revision_1}{In some cases, using the geo-coordinates got from the UAV helps to localize the boundaries of the road detected by the camera. Especially when the camera's view is orthogonal to the road itself and without any inclination. However, in many cases, these geo-coordinates could not be captured accurately for every frame of the UAV video. Moreover, these geo-coordinates become useless when the camera is inclined. In this case, deducing the correct location of the road boundaries becomes impossible without considering the images captured from the UAV. Hence, it is recommended to use computer vision for the generic approaches of road detection. } 
For example, Zhou et al. \citep{zhou2014efficient} were among the first to design an algorithm for both the detection and the tracking of roads. Moreover, the method was efficient when tested on multiple scenarios. 
Kim et al. \citep{kim2005realtime} proposed another algorithm that is trained on the road structure using one video frame. The algorithm then predicts road locations in the remaining frames of the video.

\subsection{Vehicle Detection and Tracking}
Vehicle detection and Tracking from UAV was the main target for most research works that leveraged the UAV imagery for traffic analysis \citep{hinz2005context, cheng2011vehicle, cao2011vehicle, zhao2003car, breckon2009autonomous, gaszczak2011real, kaaniche2005vision, cao2012vehicle, cao2014ego, yu2009motion, yalcin2005flow, aeschliman2014tracking, guido2016evaluating}. In addition, detecting and tracking vehicles were the primary goals since the adoption of UAVs for civil use.
Vehicle Detection is considered the first building block for any framework that analyzes traffic. Similarly, Vehicle Tracking is considered the second building block for this kind of framework. 
In the work introduced by Cao et al. \citep{cao2012vehicle}, a system of Vehicle Detection and Tracking was designed based on multi-motion layer analysis in an unsupervised way. It deduces the vehicle position by subtracting the foreground from the background. However, later, supervised algorithms were increasingly adopted due to their efficiency. Supervision given by the user enables learning the circulating vehicles' patterns using a structured dataset of UAV images. SVM and CNNs have shown significantly better efficiency in detecting vehicles \citep{cao2011vehicle, zhao2017automated}. The state-of-the-art object detectors were used to detect vehicles similarly to other objects. 

\subsection{Traffic parameters estimation}
Traffic parameters estimation from images captured using a UAV was the target of numerous research studies \citep{mccord2003estimating, angel2003methods, yamazaki2008vehicle, toth2006extracting, shastry2005airborne, ke2015motion, ke2016real, zhao2017automated, du2017open}. Their estimation could be performed offline or online, following the computation needs and the implementation constraints. Many traffic parameters were targeted in literature: density of vehicles inside the roads, speed, travel time, traffic congestion, delay due to congestion, annual average daily traffic. 
For example, McCord et al. \citep{mccord2003estimating} were among the firsts to introduce a method to estimate the annual average daily traffic from aerial imagery. Shastry et al. \citep{shastry2005airborne} tried to estimate a subset of traffic parameters based on motion information and image registration. More recently, Ke et al. \citep{ke2016real} designed a method to measure the speed, the vehicle density, and the volume of the traffic from aerial imagery. However, motion-based vehicle detection affects the method efficiency in congestive areas. Ke et al. \citep{ke2018real} tried to solve this limitation by introducing a framework that works for both congested and non-congested traffic conditions. 
\textcolor{revision_1}{Gattuso et al. \citep{gattuso2020traffic} studied the use of drones to extract traffic parameters such as: vehicle classification,  estimation of the traffic flow, the traffic density, and the speed in contexts related to urban roads and freight terminals. However, their work does not represent a solid framework for precise insights from traffic data. It represents an initial effort in this direction. Li et al.  \citep{li2021domain} targeted the problem of traffic analysis at nighttime. They proposed a domain adaptation algorithm to maximize vehicle detection accuracy at nighttime. Their method did not need a labeled dataset for nighttime images, and it is trained only on daytime images. Their work aims to introduce a  generic method for traffic parameter analysis that works on both daytime and nighttime. Ke et al. \citep{ke2020advanced} made a framework for traffic parameters estimation from moving UAV video. Although their work is interesting, the challenging problem of moving cameras makes their solution not yet mature enough to be used directly by a traffic engineer. This study aims to ensure the precision of the metrics by specifying some conditions in the UAV camera view itself. } 

\textcolor{revision_1}{Brahimi et al. \citep{brahimi2020drones} investigated the potential of the drones for traffic parameters estimation. They took as an example an urban roundabout. Then, they prove that drones are better than conventional sensors for estimating traffic parameters in this case. The study was limited and did not present a clear framework for extracting precise insights from the drone video. They intended to point out the efficiency of drones compared to conventional sensors.} 

\textcolor{revision_1}{In 2016, Khan et al. \citep{khan2017uav} made a review about the research works related to traffic analysis from UAVs. Based on it, they introduced a Framework that summarizes the technologies needed for the different steps of analyzing traffic from UAV videos. This Framework is divided into seven steps or components: (1) definition of the scope, (2) planning of the UAV flight, (3) implementation of the flight, (4) acquisition of the data, (5) processing and analysis of the data, (6) interpretation of the data, and finally (7) application for an optimized traffic. Although being valuable, their work has some limitations. First, it does not present a clear Framework for straightforward and precise extraction of measurable insights from the captured UAV video. Second, it is not focused on the UAV video analysis itself, which is the main challenging component. This component is treated in a shallow manner equally to far simpler components like the definition of the scope, for example. The reader does not have any clear insight he can precisely have from the UAV video. Moreover, the work is a bit outdated and needs to be refreshed, especially in the image analysis domain. This domain has known a considerable change in the past five years. }

\textcolor{eaai_rev_1}{From the above study, we can deduce the diversity of the research works that targeted traffic analysis from UAVs. However, to our knowledge, no one gave the community a clear and unified pipeline for the extraction of a comprehensive set of traffic data from one UAV video. Consequently, the current study is targeting this gap. It introduces a practical and straightforward pipeline for understanding the traffic from UAV video. It provides the theoretical basis and the implementation details to be followed. 
The source code is opened to the community. The Framework of this pipeline is coined as TAU (Traffic Analysis from UAVs).} The next section (Section \ref{section:proposed-method})  demonstrates the theory behind TAU before validating the implementation in Section \ref{section:experimental-results}.

\section{TAU: The Proposed Method}\label{section:proposed-method}
\subsection{Method Overview}
The TAU framework is based on two main building blocks: \textit{Vehicle Detection} and \textit{Vehicle Tracking}. For the first building block, the vehicle detection task, it operates on the UAV video frame by frame to detect all the occurrences of vehicles inside one frame. To this end, we built a specific vehicle detection dataset on UAV images. The annotated vehicles inside the dataset were grouped into five classes (Car, Bus, Truck, Bicycle/Motorcycle, Pedestrian). Then, we trained one state of the art object detector (YOLO v3 \citep{YOLOv3}) on this dataset to detect vehicles inside the UAV video. \par
\textcolor{eaai_rev_1}{Concerning the choice of YOLO v3, it can be replaced by another object detector: either a generic object detector or a dedicated object detector for aerial imagery. For generic object detectors, we can choose YOLO v4 \citep{bochkovskiy2020yolov4} which is a more recent version of the YOLO family released in April 2020. It comes with many improvements over YOLO v3. However, we used YOLO v3 in TAU because, at the time of the implementation of the code, YOLO v4 was not released. The replacement of YOLO v3 by YOLO v4 is one of the improvements we listed for the next version of TAU (see Section V). Moreover, a dedicated object detector for aerial imagery can be used. One of the most recent works in this field is the Butterfly Detector \citep{adaimi2020perceiving}. }\par
Although the model is trained to detect pedestrians, this class is omitted during the study to focus only on vehicles in the current stage. Later, if there is a need to analyze the pedestrian movements, the same model can be customized. The trained model extracts for every frame the bounding boxes around every vehicle. This information is saved to use later by the second building block.\par
Concerning the second building block, the vehicle tracking task, it takes as input the detected vehicles in the current frame and tries to associate every one of them to their correct ID. There are two cases for this association.
In the first case, if one of the detected vehicles has not been seen before in the previous frames, the tracking algorithm associates to it a new identifier. In the second case, if this newly detected vehicle existed already in one of the previous frames and got an identifier \(X\), the tracking algorithm associates this vehicle to this identifier. After that, all the bounding boxes related to one vehicle during the video are saved in a separate dataset. These data help generate the statistics related to the vehicle movement over the whole video. \textcolor{eaai_rev_1}{Concerning the tracking algorithm selected for TAU, we picked out one state of the art algorithm in the domain of Real-Time Multi-Object Tracking: DeepSORT       \citep{deep_sort}.}\par
In the following subsections, we will discuss the theoretical basis of the two building blocks of TAU (Subsections \ref{section:yolov3} and \ref{section:deep_sort}). Next, the theoretical basis of the different \textcolor{revision_1}{insights} will be discussed based on the dataset generated at the end of the two building blocks. 

\subsection{Vehicle detection using YOLO v3}\label{section:yolov3}
\subsubsection{YOLOv3 overview}
Introduced in 2018, YOLO v3 \citep{YOLOv3} is an incremental improvement made over the previous versions: YOLO v1 \citep{YOLO2016} introduced in 2016 and YOLO v2 \citep{YOLOv2} in late 2016. It is a popular object detection algorithm, especially for real-time applications due to its efficient processing time compared to competitors \citep{YOLOv3}. \par 

\begin{figure*}  
\begin{center}  
\includegraphics[width=18cm]{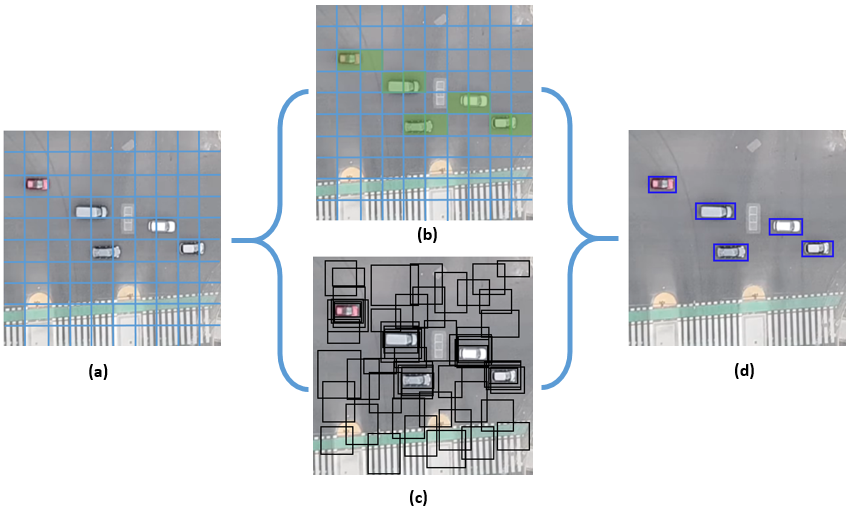}
\caption{\small \sl The detection process of YOLO v3. (a) input image is divided into a grid of \(13*13\) cells, every cell is of size \(32*32\) pixels. (b) The bounding boxes with their confidence score are predicted for every cell. (c) Generation of the class probability for every grid cell. (d) final bounding box with class association are finally deduced. 
\label{fig:yolo_process}}  
\end{center}  
\end{figure*}

The YOLO v3 detection process is clarified more in \(Figure\) \ref{fig:yolo_process}. First, the input image is resized to \(416*416\). The resulting image is divided into a grid of \(13*13\) cells; every cell is of size \(32*32\) pixels. In YOLO v3, the number of the grids depends on the size of the input image. In the current case, the input image given to the network is of size \(416*416\), which can be divided into a grid of \(13*13\) cells, every cell is of size \(32*32\). If it was \(608*608\), it will be divided into a \(19*19\) grid, every cell is always of size \(32*32\). After the partitioning step, the model predicts the class probability and the bounding boxes for every grid cell with a confidence score. This information is assembled to deduce the final detected object in the input image (most relevant bounding box and the class it belongs to).

\begin{figure*}  
\begin{center}  
\includegraphics[width=18cm]{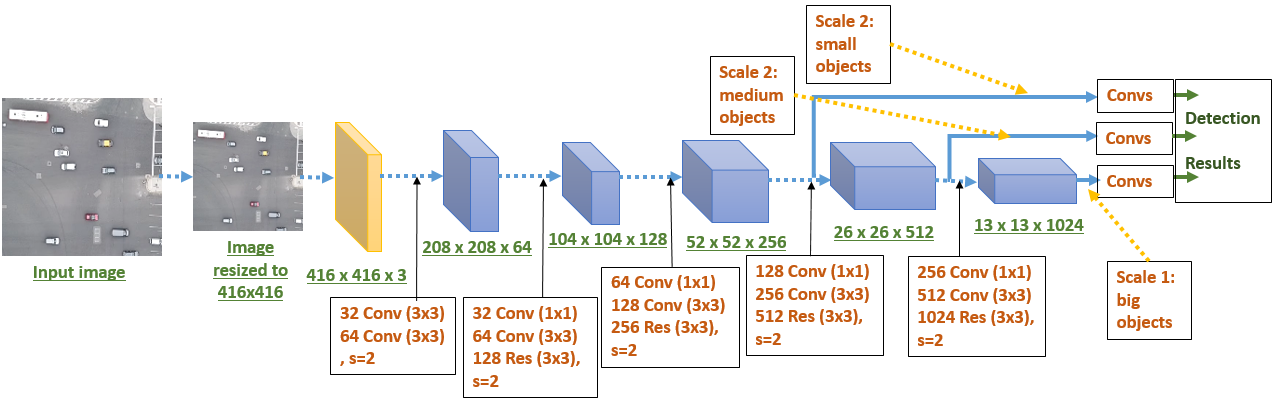}
\caption{\small \sl Network architecture of YOLO v3. 
\label{fig:yolo_architecture}}  
\end{center}  
\end{figure*}

The model is trained on six classes: class Car, class Bus, class Truck, class Motorcycle or Bicycle, and finally the class Pedestrian (this class is omitted in this study). YOLO v3 does the whole prediction using only one convolutional network. It takes as input the resized image and predicts as output a tensor of dimension 
\begin{equation} \label{eq:6}
 N\times N\times (3*(5+C))  
\end{equation}
where:
\begin{itemize}
\item \( N\times N\): is the number of grid cells.
\item \(C\): is the number of targeted classes.
\item \((3*(5+C))\): This is because 5 parameters are detected (\( t_x, t_y, t_w, t_h \), and the box confidence score) plus the probability for every one of the classes (\(C\)). This is done for every grid cell at 3 different scales. 
\end{itemize}
In the trained model, this tensor will be of size \(13*13*(3*(5+6)\),  which corresponds to a 3D tensor of dimension \(13*13*33\).
The main architecture of YOLO v3 network is displayed in \(Figure\) \ref{fig:yolo_architecture}. 

The UAV collected the images from an altitude of 175 meters. This altitude is chosen to maximize the surveyed area. This gives a wide perception of a crossroad intersection. Including a clear view of the roads carrying vehicles inside and outside the intersection. The video is captured in 4K resolution (\(4096 * 2160 \) pixels). High resolutions like 4K are recommended for better analysis of the traffic. Lower resolutions could lead to blurred views or lower accuracy. \par 

\subsubsection{Dealing with High-Resolution Images}
Although high resolutions are recommended for the input video, they cannot be used directly in the training of YOLO v3. This limitation is due to the heavy computational load. The maximum resolution that can be used at a reasonable computation cost is \(608*608\). For the construction of the training dataset, a perpendicular view on the ground should always be kept. Therefore, only UAV images that are perpendicular to the ground are considered in this study. This constraint simplifies one dimension in the mathematical equations used during the analytics. For the equations used in the \textcolor{revision_1}{insights'} extraction, the following hypothesis is assumed: The vehicles are circulating on the plane of the equation:
\begin{equation} \label{eq:hyp_z_eq_0}
 [z = 0], (x, y, z) \in \Re 
\end{equation}
Where \((z)\) refers to the third dimension in the 3D coordinate system  \(\Re(x,y,z)\). More details about the mathematical formulation are given in section \ref{section:knowledge_extraction}\par  
Finally, the dataset is constructed by collecting croppings of size \(512*512\) from different locations inside the captured UAV images, but croppings that contain vehicles are preferred. After that, the annotation of the different classes is made.\par

During the framework test on a sample UAV video of the traffic, the view of the test video should be perpendicular to the ground. 
During the execution of the TAU framework, the test UAV video is processed frame by frame. Every frame dimension is extended to the nearest integer dividable by \(512\), as shown in Algorithm \ref{preprocessing_algo}.
\begin{algorithm}[h]
\label{preprocessing_algo}
\SetKwInOut{Input}{input}\SetKwInOut{Output}{output}
\Input{$frame$: the input frame}
\Output{$preprocessed\_frame$: the pre-processed frame}
\BlankLine
\SetAlgoLined
$crop\_height$ = 512;\\
$crop\_width$ = 512;\\
$height$ = height($frame$);\\
$width$ = width($frame$);\\
  \If{($height$ mod $crop\_height$ != 0)}{
   $height$ = $crop\_height$ * ($height$ div $crop\_height$) +1;\\
   }
  \If{($width$ mod $crop\_width$ != 0)}{
   $width$ = $crop\_width$ * ($width$ div $crop\_width$) +1;\\
   } 
   $preprocessed\_frame$ = Array($width$, $height$, 3);\\
   \For{($i$=0; $i$ $<$ $width$; $i$++)}{
   \For{($j$=0; $j$ $<$ $height$; $j$++)}{
  \eIf{($frame$[i,j])}{
   $preprocessed\_frame$[i,j] = $frame$[i,j];\\
   }{
   $preprocessed\_frame$[i,j] = [0,0,0];\\
   }      
   }   
   }
 \caption{Pre-processing the input frame}
\end{algorithm}

After preparing the pre-processed frame, it is divided into a grid of processed cells. Every processed cell is of size \(512*512\) pixels. Each of them is then passed to the YOLO v3 trained model to detect the vehicles inside it. Then, the resulting local coordinates for every vehicle are scaled accordingly to deduce their coordinates in the global frame. This task is done following the Algorithm \ref{algo-detect-vehicles}.

\begin{algorithm}[!h]
\SetKwInOut{Input}{input}\SetKwInOut{Output}{output}
\Input{$preprocessed\_frame$: the pre-processed frame}
\Output{$detected\_vehicles$: the detected vehicle in the input frame}
\BlankLine
\SetAlgoLined
$crop\_height$ = 512;\\
$crop\_width$ = 512;\\
$W$ = width($preprocessed\_frame$);\\
$H$ = height($preprocessed\_frame$);\\
define struct Vehicle \{int box\_x ; int box\_y; int box\_width; int box\_height; int class\};

$detected\_vehicles$ = List();\\
$local\_detected\_vehicles$ = List();\\
$processed\_cell$ = Array($crop\_width$, $crop\_height$, 3);\\
\For{($div\_width$=0; $div\_width$ $<$ W; $div\_width$++)}{
\For{($div\_height$=0; $div\_height$ $<$ H; $div\_height$++)}{
\For{($i$=0; $i$ $<$ $crop\_width$; $i$++)}{
\For{($j$=0; $j$ $<$ $crop\_height$; $j$++)}{
$processed\_cell$[i,j] = $preprocessed\_frame$[\\i+$crop\_width$*$div\_width$,\\ j+$crop\_height$*$div\_height$]

}
}
$local\_detected\_vehicles$ =\\ YOLOV3($processed\_cell$);\\
\lForEach{Vehicle $v$ in $local\_detected\_vehicles$}
{
\\
$detected\_vehicles$.add(new Vehicle(\\$v$.box\_x +$crop\_width$*$div\_width$\\,
$v$.box\_y +$crop\_height$*$div\_height$\\, $v$.box\_width, $v$.box\_height, $v$.class
))
}
}
}
\caption{Detecting vehicles in the input frame}\label{algo-detect-vehicles}
\end{algorithm}
In the end, the final list of the detected vehicles are passed as input for the tracking algorithm to decide for every vehicle if it belongs to an already tracked vehicle, or it is a newly intervening car inside the scene (as detailed in Section \ref{section:deep_sort}). 

\subsection{Vehicle tracking using Deep SORT algorithm}\label{section:deep_sort}

\begin{figure*}[h]  
\begin{center}  
\includegraphics[width=18cm]{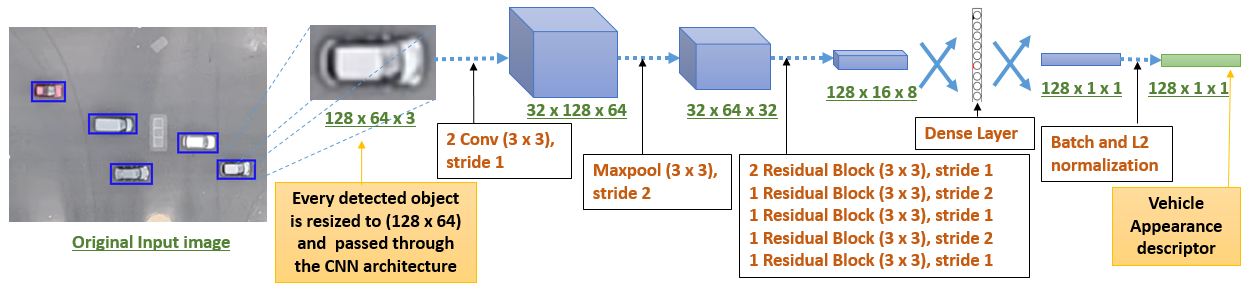}
\caption{\small \sl Generation of appearance descriptor for every detected object.
\label{fig:appearance_descriptor}}  
\end{center}  
\end{figure*}

To track the detected vehicles, the DeepSORT algorithm \citep{deep_sort} is chosen. It is a state-of-the-art online multi-object tracking algorithm \citep{ciaparrone2020deep}. 
It was an improvement over its earlier version: the SORT \citep{tracking_sort} (Simple Online and Real-time Tracking) algorithm. DeepSORT has an accuracy comparable to the state-of-the-art trackers but with a better processing time (up to 60 frames per second). \par 

In TAU, DeepSORT processes the UAV video frame by frame. For every frame, it takes the detected vehicles generated by the Algorithm \ref{algo-detect-vehicles}. Then, it tries to associate them to the previously tracked vehicle ID or a new tracked vehicle ID. Finally, the motion model state of the vehicle is implemented using eight parameters:
\begin{equation} \label{eq:9}
\textcolor{eaai_rev_1}{X= [x, y, a, h, \dot{x}, \dot{y}, \dot{a}, \dot{h}]}
\end{equation}
where \(x\) and \(y\)  are respectively the horizontal and the vertical coordinates of the vehicle location in the image, \textcolor{eaai_rev_1}{which is associated to the center of its bounding box, \(a\) is the aspect ratio of the vehicle bounding box, \(h\) is the height of the bounding box, and the other parameters are their respective estimated velocities in the image domain.} Kalman filter is used to predict the new motion state of every vehicle based on the previous measurements. \par
If a detected vehicle is associated to an ID, the bounding box parameters are used to update the target motion model state parameters using a Kalman filter. The association of the newly detected objects to the already tracked targets is made via calculating the IoU (Intersection Over Union) between  every detection bounding box and all the Kalman-filter predicted bounding boxes of the existing vehicle IDs. If the IoU is lesser than a fixed value \(IoU_{Min}\), the assignment is rejected and the detected object is assigned to a new target. The assignment is implemented via the Hungarian algorithm. \par
Deep SORT \citep{deep_sort} added the appearance information into the association operation. A designed Convolutional Neural Network (CNN) architecture generates a deep appearance descriptor for every detected vehicle. The CNN architecture is trained offline on a large re-identification dataset.The CNN architecture is illustrated in \(Figure\) \ref{fig:appearance_descriptor}. The inclusion of the appearance descriptor in DeepSORT improved the algorithm's ability to deal with longer periods of occlusions without affecting the high frame rate during the processing. This descriptor has also reduced the number of identity switches. Besides, it made the algorithm robust to the possible overlaps between the bounding boxes of different tracked vehicles. 

Two metrics are combined to associate the new detected vehicles to the tracked targets. The first metric is the motion metric. It is expressed as the squared Mahalanobis distance between the detected object motion state and the predicted motion state for every tracked target already saved. The squared Mahalanobis distance is defined in \(Equation\) \ref{eq:10}: 
\begin{equation} \label{eq:10}
d^{(1)}(i,j) = (d_j - y_i)^T S^{-1}_i (d_j - y_i)
\end{equation}
where \(d_j\) is the \(j^{th}\) detected bounding box, \((y_i, S_i)\) is the projection of the \(i^{th}\) track distribution into the measurement space. \par

The second metric is the smallest cosine distance between the appearance descriptor of the detected object \(r_j\) and the previous appearance descriptors of the \(i^{th}\) tracked vehicle ID in the previous frames . This metric is expressed in \(Equation\) \ref{eq:11}:
\begin{equation} \label{eq:11}
d^{(2)}(i,j) = min\{1-r^T_j r^{(i)}_k | r^{(i)}_k \in R_i\} 
\end{equation}
where \(R_i\) is the list of the last appearance descriptors for the \(i^{th}\) tracked vehicle, \(R_i = \{r^{(i)}_k\}_{k=1}^{L_k}\), \textcolor{eaai_rev_1}{\(k\) is the index of the appearance descriptor belonging to this list \(R_i\). This \(k\) is an integer that increments from 1 to a limit value \(L_k\) that is a constant set by the user during the implementation. In our case, \(L_k\) took the value 100 during the experiments.} \par
The two metrics are combined using a weighted sum:
\begin{equation} \label{eq:12}
c_{i,j} = \lambda d^{(1)}(i,j) + (1-\lambda)d^{(2)}(i,j)  
\end{equation}
The \(\lambda\) parameter controls the weight of every metric inside the association. 

After the association step, every detected vehicle inside the current frame will be associated with a vehicle ID. If the vehicle is already present in the previous frames and got a specific ID, it will be assigned to this ID. If the vehicle is newly seen inside the scene, it will be assigned to a new ID. This ID is pivotal information in the TAU framework. Most statistics calculated in the next subsection are  based on it (please see subsection \ref{section:knowledge_extraction}).  
\subsection{\textcolor{revision_1}{Extraction of insights} from the UAV video}\label{section:knowledge_extraction}

After processing the frame, the system generates a list of the detected vehicles inside this frame. Two major pieces of information characterize every vehicle. The first is the bounding box information (Generated by Algorithm \ref{algo-detect-vehicles}). The second is the tracked ID of the vehicle; this is assigned by Equation \ref{eq:12}. For every processed frame ID, a separate database of statistics is needed to analyze the movement of the vehicles. The format of the relational table used to record these statistics is described in \(Table\) \ref{tab:table_format}.\par 

\begin{table}[h!]
\centering
\resizebox{9cm}{!}{
\begin{tabular}{lll}
\hline
\textbf{\begin{tabular}[c]{@{}l@{}}Column\\ Number\end{tabular}} & \textbf{\begin{tabular}[c]{@{}l@{}}Column \\ Name\end{tabular}}                & \textbf{Description}                                                                                                                                                                                                                                                     \\ \hline
\textbf{1}                                                       & \textbf{Index}                                                                 & \begin{tabular}[c]{@{}l@{}}The Primary Key of the table, \\ an Integer value that \\ increments with every new \\ row added inside the table. \\ Every row corresponds to \\ a separate recording.\end{tabular}                                                          \\ \hline
\textbf{2}                                                       & \textbf{Frame ID}                                                              & \begin{tabular}[c]{@{}l@{}}The frame ID already \\ processed in the current\\ recording.\end{tabular}                                                                                                                                                                    \\ \hline
\textbf{3}                                                       & \textbf{\begin{tabular}[c]{@{}l@{}}Bounding box \\ coordinates\end{tabular}}   & \begin{tabular}[c]{@{}l@{}}The bounding box coordinates\\  of the detected vehicle \\ considered in this recording. \\ The bounding boxes of all the \\ detected vehicles are generated\\  by the Algorithm \ref{algo-detect-vehicles}.\end{tabular} \\ \hline
\textbf{4}                                                       & \textbf{Vehicle ID}                                                            & \begin{tabular}[c]{@{}l@{}}The tracked ID assigned by \\ DeepSORT to the detected \\ vehicle. The ID is assigned \\ based on the Association \\ metric defined in the \\ Equation \ref{eq:12}.\end{tabular}                                              \\ \hline
\textbf{5}                                                       & \textbf{\begin{tabular}[c]{@{}l@{}}Exact time \\ (seconds)\end{tabular}}       & \begin{tabular}[c]{@{}l@{}}The exact time \\ corresponding to the current \\ frame in seconds.\end{tabular}                                                                                                                                                              \\ \hline
\textbf{6}                                                       & \textbf{\begin{tabular}[c]{@{}l@{}}center of the \\ bounding box\end{tabular}} & \begin{tabular}[c]{@{}l@{}}The center of the bounding \\ box corresponding to the \\ detected vehicle considered \\ in the current recording.\end{tabular}                                                                                                               \\ \hline
\textbf{7}                                                       & \textbf{\begin{tabular}[c]{@{}l@{}}size of the \\ vehicle\end{tabular}}        & \begin{tabular}[c]{@{}l@{}}The approximative measure\\  of the size of the vehicle: \\ the area of the bounding box.\end{tabular}                                                                                                                                        \\ \hline
\textbf{8}                                                       & \textbf{\begin{tabular}[c]{@{}l@{}}Velocity in \\ km per hours\end{tabular}}   & \begin{tabular}[c]{@{}l@{}}The instantaneous velocity \\ of the vehicle during the \\ exact time of the current \\ frame.\end{tabular}                                                                                                                                   \\ \hline
\textbf{9}                                                       & \textbf{Zone}                                                                  & \begin{tabular}[c]{@{}l@{}}The zone where the \\ vehicle is located. It will \\ associate the value 0 if it is \\ not inside the user-defined \\ zones. (7 zones are defined\\ in the test).\end{tabular}                                                                \\ \hline
\end{tabular}}
\caption{\small \sl Format of the Data stored to record the vehicle movements and statistics over time.
\label{tab:table_format}} 
\end{table}
The considered relational table contains nine columns. The first column is the index, an integer value that increments with every recording added to the table. Then, we have the three primary columns already described above: the Frame ID, the Bounding box coordinates, and the Vehicle ID. From these three columns, we automatically deduce the remaining five columns. These five columns can be calculated online during the processing of the frame or offline after the processing of the total UAV video. They are not extracted directly from the video like the three main column values. \par
Hence, at the end of processing one frame, the data related to every vehicle is saved in separate rows. Every row corresponds to the information of every vehicle detected in the processed frame.\par
From the saved database, a set of traffic information could be extracted. The theoretical basis for every extracted information is detailed inside the following paragraphs.
\subsubsection{\textbf{Exact time of the frame}}
This information can be directly deduced from the Column 2 (Frame ID) by using the following Equation: 
\begin{equation} \label{eq:k01}
ET = \frac{FID}{FPS}
\end{equation}
Where \(ET\) is the Exact time expressed in seconds. \(FID\) is the Frame ID. It is an integer value initialized to zero before processing the video. It is incremented after the processing of every frame apart. \(FPS\) is the frame per second information extracted directly from the UAV video. 
\subsubsection{\textbf{Center of the bounding box of the vehicle}}
This information can be extracted directly from Column 3 (Bounding box coordinates of the Vehicle). The Bounding box can have two possible formats. The first format is: 
\begin{equation} \label{eq:k02}
[x_0, y_0, x_1, y_1]
\end{equation}
Where \((x_0, y_0)\) are the coordinates of the upper left corner of the Bounding Box, \((x_1, y_1)\) are the coordinates of the lower right corner of the Bounding box. For this format, the center point is deduced using the following Equation: 
\begin{equation} \label{eq:k03}
C(x_c, y_c) = (\frac{x_0+x_1}{2}, \frac{y_0+y_1}{2})
\end{equation}
The second format for the bounding box is: 
\begin{equation} \label{eq:k04}
[x_0, y_0, w, h]
\end{equation}
Where \(w\) and \(h\) are respectively the width and the height of the bounding box. For this format, the center point is deduced using the following Equation:
\begin{equation} \label{eq:k05}
C(x_c, y_c) = (x_0 +\frac{w}{2}, y_0 + \frac{h}{2})
\end{equation}
\subsubsection{\textbf{Approximate size of the vehicle}}
Following the Hypothesis defined in the Equation \ref{eq:hyp_z_eq_0}, the size of the top-down view of the vehicle can be approximated to the Area of the bounding box. Since the bounding box area is expressed in pixels, we need to convert it to the metric scale using the UAV image's Ground Sample Distance \(GSD\). \(GSD\) is expressed using the following Equations:
\begin{equation} \label{eq:k06}
GSD_h = \frac{DAG * S_h}{FL * Im_h}
\end{equation}
\begin{equation} \label{eq:k07}
GSD_w = \frac{DAG * S_w}{FL * Im_w}
\end{equation}

where \(DAG\) is the distance above the ground, it corresponds to the height between the lens of the UAV camera and the ground. \(S_h\) and \(S_w\) are respectively the height and the width of the camera sensor itself. \(FL\) is the focal length of the camera, which is the distance between the sensor of the camera and its lens. All of \(DAG, S_h, S_w\) and \(FL\) are expressed in meters. \(Im_h\) and \(Im_w\) are the width and the height of the whole captured image in pixels. Finally \(GSD_h\) and \(GSD_w\) are expressed in meter per pixel \(m/px\).\par
From \(GSD_h\) and \(GSD_w\), we can deduce the Ground  Sample Area \(GSA\), which is the area associated with every pixel in the UAV image. It is expressed in squared meter  per pixel \(m^2/px\) and calculated using the following Equation:
\begin{equation} \label{eq:k08}
GSA = GSD_h * GSD_w
\end{equation}
Based on \(GSA\), we can calculate the size of the vehicle \(A\), which is approximately the area of the Bounding box. \(A\) is expressed in squared meters \(m^2\) and calculated using the following Equation: 
\begin{equation} \label{eq:k09}
A = (x_1 - x_0) * (y_1 - y_0) * GSA 
\end{equation}
This Equation is valid for the format defined in the Equation: \ref{eq:k02}. However, for the second format defined in the Equation \ref{eq:k04}, It is expressed as:
\begin{equation} \label{eq:k10}
A = w * h * GSA
\end{equation}

\subsubsection{\textbf{Vehicle Velocity}}

During the video processing, the instant velocity of the tracked vehicle is updated for every frame. It is initialized to \(0 \  km/h\) for the first frame. Then, it is updated by deducing it from the database referred in \(Table\) \ref{tab:table_format}. However, five hypotheses are assumed to simplify the velocity estimation based on the constraints predefined in this study. \par 
First, we consider the Cartesian coordinate system:
\begin{equation} \label{eq:k11}
\Re(O,\vec{x},\vec{y},\vec{z})
\end{equation}
\(O\) is the origin of the coordinate system located on the physical point on the ground surface projected on the upper left corner of the UAV image. This point is fixed during all the processed UAV video. \(\vec{x}\), \(\vec{y}\) and \(\vec{z}\) are the unitary vectors of this coordinate system: 
\begin{equation} \label{eq:k12}
\Vert \vec{x} \Vert = \Vert \vec{y} \Vert =  \Vert \vec{z} \Vert = 1 m 
\end{equation}
They are also orthogonal to each other: 
\begin{equation} \label{eq:k13}
\vec{x} \perp \vec{y} \perp \vec{z} 
\end{equation}
\(\vec{x}\) is oriented from the origin \(O\) to the physical point on the ground surface projected on the upper right corner of the UAV image. Similarly, \(\vec{y}\) is oriented from the origin \(O\) to the physical point on the ground surface projected on the lower-left corner of the UAV image. \(\vec{z}\) is orthogonal to the plane \((\vec{x}, \vec{y}) \) and oriented from the origin \(O\) following the right-hand rule. \par
Second, we assume that the physical ground surface is approximately a part of  the hyperplane \(P_0\) of \(\Re\) defined by:
\begin{equation} \label{eq:k14}
P_0: (x, y, z = 0)
\end{equation}
This hypothesis engenders that the z component of the vehicle velocity is null. Hence, the velocity of any vehicle on UAV video can be expressed as:
\begin{equation} \label{eq:k15}
\vec{V} = V_x.\vec{x} + V_y.\vec{y}
\end{equation}

Third, we assume that the camera view is perpendicular to the ground surface following the same direction of \(\vec{z}\). Hence, the surface of the camera sensor is a part of the hyperplane \(P_1\) defined by:
\begin{equation} \label{eq:k16}
P_1: (x, y, z = DAG + FL)
\end{equation}
Thus, we deduce that the hyperplanes \(P_0\) and \(P_1\) are parallel:
\begin{equation} \label{eq:k17}
P_0 \Vert P_1
\end{equation}
\textcolor{eaai_rev_1}{Now, we model the camera following the standard pinhole camera model. The camera will have its separate coordinate system \(\Re(C,\vec{i},\vec{j},\vec{k})\), where \(C\) is the center of the camera, which corresponds to the pinhole itself. \(\vec{i}\) is parallel to \(\vec{x}\) and have the same norm and distance. Also,  \(\vec{j}\) is parallel to \(\vec{y}\) and have the same norm and distance. \(\vec{k}\) is perpendicular to the sensor and points toward it. Every point \(P_i\) seen on the image is a projection of a real world point \(P_w\) that exists on the ground. The projection follows the camera matrix model presented by the equation \ref{eq:camera_model}:
\begin{equation} \label{eq:camera_model}
P_i = M*P_w = K*[R\ \ T]*P_w
\end{equation}
where \(M\) is the full projection matrix of the camera. \(M\) combines both the intrinsic and extrinsic parameters. The intrinsic parameters are identified by the matrix \(K\). The extrinsic parameters are represented the matrix \([R\ \ T]\). \(R\) and \(T\) are the rotation and the translation matrices between the world coordinate system \(\Re(O,\vec{x},\vec{y},\vec{z})\) and the camera coordinate system \(\Re(C,\vec{i},\vec{j},\vec{k})\).  
} \par
\textcolor{eaai_rev_1}{Based on the previous hypotheses, the rotation matrix \(R\) is null. Also, the translation matrix has only one component over the axis \(\vec{z}\). Based on the equation \ref{eq:k14}, all the components  over \(\vec{z}\) can be neglected and we can work only on \(\vec{x}\) and \(\vec{y}\). Thus, we can omit the extrinsic parameters of the camera \([R\ \ T]\) and keep only  the intrinsic parameters \(K\). After calibration of the camera, these intrinsic parameters are identified which permits to easily have the projection coordinates of every seen point on the ground from the camera perspective without need of conversion between the coordinate systems. Hence, we can assume for our case the following camera projection model identified in the equation \ref{eq:simplified_model} below:
\begin{equation} \label{eq:simplified_model}
P_i = 
\begin{bmatrix}
x_i \\ y_i
\end{bmatrix}
= K*P_w = K* \begin{bmatrix}
x_w \\ y_w
\end{bmatrix}
\end{equation}
} 
\textcolor{eaai_rev_1}{After that, it remains only to convert the projected  image  from the pixel scale into the metric scale using the Equations \ref{eq:k06}, \ref{eq:k07} and \ref{eq:k08}. Hence, we can follow our calculations on the  2D representation of the vehicle displayed on the UAV image converted to the metric scale. It is an exact camera projection of the corresponding vehicle on the ground surface belonging to the hyperplane \(P_0\) (Equation \ref{eq:k14}).  Moreover, based on  the Equation \ref{eq:k15}, we can deduce that the Velocity \(\vec{V}\) of a vehicle is the same as the velocity of the 2D projection of the vehicle displayed on the UAV image converted to the metric scale.
} \par

The fourth hypothesis to consider is that the inertial point needed to estimate the global velocity of the vehicle can be approximated as the center point of the 2D bounding box of the vehicle. This approximation is based on the previous equations and deductions.\par
The fifth hypothesis to consider is that the movement of one vehicle from one point \(C_{(n)} (x_c^{(n)}, y_c^{(n)})\) (the center point of the bounding of the vehicle in the frame \((n)\)) to the point \(C_{(n+1)} ((x_c^{(n+1)}, y_c^{(n+1)}))\) (the center point of the bounding of the vehicle in the frame \((n+1)\)) can be approximated as a linear movement from \(C_{(n)}\) to \(C_{(n+1)}\). This allows to Approximate the instant velocity between these two frames as the Euclidian distance between \(C_{(n)}\) an \(C_{(n+1)}\) divided by the time \(\Delta t = \) needed for this movement. This time corresponds to the time between two frames, which is:
\begin{equation} \label{eq:k18}
\Delta t =  (t^{(n+1)} - t^{(n)}) = \frac{1}{FPS}
\end{equation}
where \(FPS\) is the frame rate of the UAV video, \(\Delta t\) is expressed in seconds. \par
The Euclidian distance \(D\) between \(C_{(n)}\) and \(C_{(n+1)}\) can be expressed as:  
\begin{equation} \label{eq:k19}
\begin{cases}
\alpha = (x_c^{(n+1)} - x_c^{(n)})*GSD_w \\
\beta = (y_c^{(n+1)} - y_c^{(n)})*GSD_h \\
D = \sqrt{ \alpha^2 + \beta^2 }
\end{cases}
\end{equation}
where \(\alpha\) is the movement of the vehicle following the axis \(\vec{x}\) and \(\beta\) is the movement of the vehicle following the axis \(\vec{y}\). All of \(\alpha\), \(\beta\) and \(D\) are estimated in the metric scale. 
From all the equations above, the velocity of one vehicle \(V\) can be expressed in the metric scale \(m/s\) or  \(km/h\) using the following Equations:

\begin{equation} \label{eq:k19}
V (m/s) = D * FPS 
\end{equation}

\begin{equation} \label{eq:k20}
V (km/h) = 3.6 * D * FPS
\end{equation}
\subsubsection{\textbf{Number of vehicles per zone}}

To give better analytics on the different portions of the roads, the user can freely define a set of zones to generate statistics for every zone apart. A zone is a closed polygon composed of points in a specific order. Every closed polygon starts from one point and ends by the same point forming a cycle. Every two successive points are connected using a segment. The set of points is named the vertices of the polygon. The list of segments is named the edges of the polygon. We consider \(p\) the set of zones defined by the user. By default, \(Z_0\) is considered the default zone outside the zones defined by the user. For \((i = 1..p)\), Every defined zone by the user \(Z_i\) can be formulated as: 
\begin{equation} \label{eq:k21}
Z_i = \{(x_0^{(i)}, y_0^{(i)}), (x_1^{(i)}, y_1^{(i)}),..., (x_{n_i}^{(i)}, y_{n_i}^{(i)})\} 
\end{equation}
\(Z_i\) is a closed polygon, which has the following constraints:
\begin{equation} \label{eq:k22}
(x_0^{(i)}, y_0^{(i)}) = (x_{n_i}^{(i)},..., y_{n_i}^{(i)})
\end{equation}
For every frame, every vehicle is assigned to the correct zone: user-defined zones \(\{Z_i, i=1..p\}\) or the default zone \(Z_0\). To be able to do this assignment, we adopt the approximation that if the center point \((x_c, y_c)\) of the vehicle is inside one user-defined zone \(Z_i\), it will be assigned to it. Otherwise, it will be assigned to zone \(Z_0\). \par
The problem of checking whether \((x_c, y_c)\) is inside the closed polygon \(Z_i\) is named in computational geometry the PIP problem, PIP means Point In Polygon. The Ray Casting algorithm is among the most used algorithms to solve this problem. We adopted it for counting vehicles inside the zones. The Ray Casting is based on two steps. The first step is to draw a ray that starts from \((x_c, y_c)\) and follows any direction. The second step is to count the number of times this ray intersects with the polygon edges \(Z_i\). If this number is odd, then  \((x_c, y_c)\) is inside \(Z_i\). Otherwise, it is outside it. \textcolor{eaai_rev_1}{For every frame, we loop over all the tracked vehicles to associate every one of them to the right zone. The process is formulated in Algorithm \ref{ray-tracing}}.

\begin{algorithm}[h]

\SetKwInOut{Input}{input}\SetKwInOut{Output}{output}
\Input{
$V = \{V_i =[id_i,x_i, y_i, w_i, h_i]; 0<i<N_v\}$: list of the tracked vehicles \\
$Z = \{Z_j; 1<j<N_z\}$: list of the traffic zones}
\Output{$T = \{T_i =[id_i,z_i]; 0<i<N_v\}$: list of vehicles and their corresponding traffic zones}
\BlankLine
\SetAlgoLined
\ForEach{$V_i$ in $V$}{
\# calculate the center of the vehicle 
$C=(\frac{2*x_i + w_i}{2}, \frac{2*y_i + h_i}{2})$  \\
\ForEach{$Z_j$ in $Z$}{
$is\_inside$ = False \\ 
$n$ = 0 \# Number of Intersections \\
\ForEach{$edge$ in $Z_j$}{
\If{($ray\_intersects\_segment$($C$, $edge$)}
{
$n$ =  $n$ + 1 ; 
}}
\If{($ODD$($number\_of\_intersections$)} 
{
$is\_inside$ = True \\
T.add([$V_i[0],j]$)
}
}
\# Otherwise $V_i$ belongs to the default zone $Z_0$
T.add([$V_i[0],0]$)
}
\caption{Association of the vehicles to their traffic zones}\label{ray-tracing}
\end{algorithm}

\begin{figure*}[!h]  
\begin{center}  
\includegraphics[width=18cm]{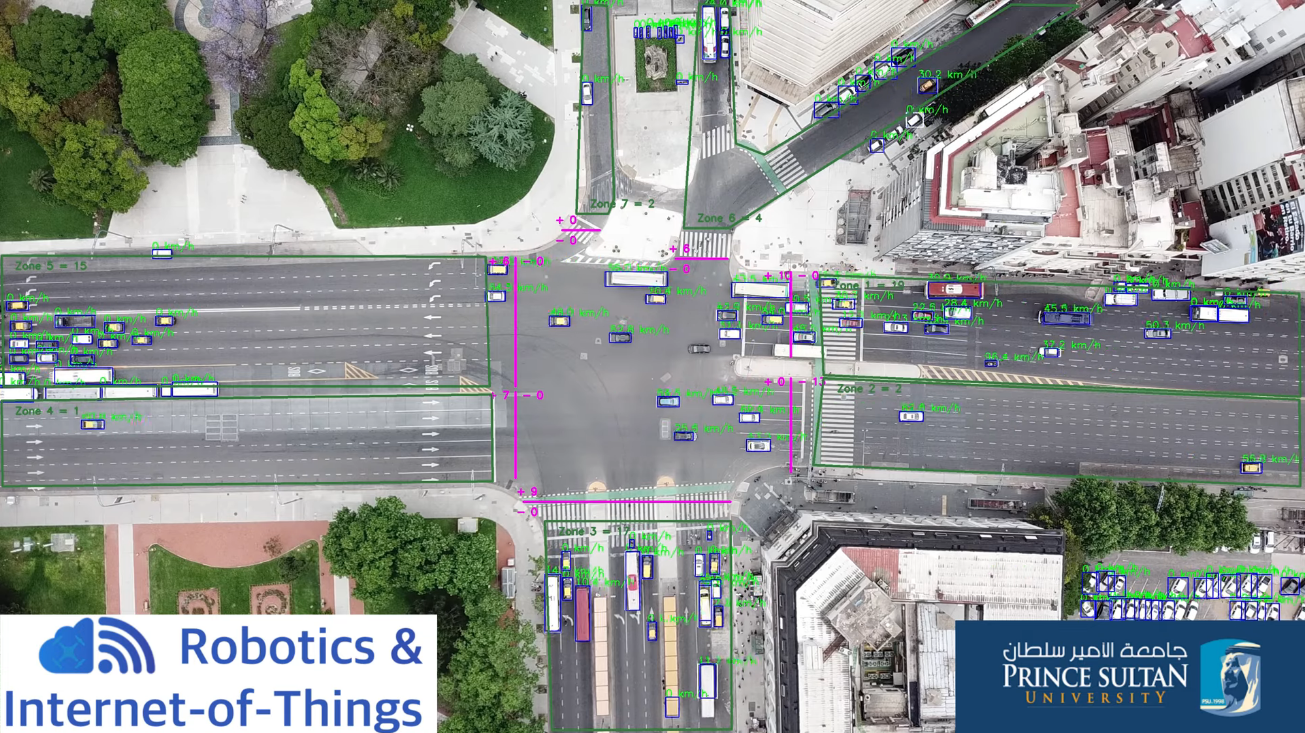}
\caption{\small \sl The 7 main types of \textcolor{revision_1}{insights} extracted using TAU from the UAV video: Vehicles bounding boxes coordinates, Vehicles ID, Vehicles Velocities, Counting Vehicles crossing specific lines from both directions, Vehicle counting in specific zones, Vehicles center points.  
\label{fig:main_knowledge_types}}  
\end{center}  
\end{figure*} 
\subsubsection{\textbf{Other extracted types of \textcolor{revision_1}{insights}}}
From the above,  TAU is based on seven essential types of \textcolor{revision_1}{insights} extracted from the UAV video: the Bounding Box Coordinates, The Vehicle ID, the Exact time of the frame, the Center of the Bounding Box, the size of the vehicle, the Velocity, and the Zone. In this section, the mathematical formulation of these types of \textcolor{revision_1}{insights} is deeply detailed. Based on them, the Transport Engineer can extract many other types of \textcolor{revision_1}{insights} to analyze the traffic information efficiently. In the Experimental results section (Section \ref{section:experimental-results}), TAU is applied to generate 17 other types of \textcolor{revision_1}{insights}. Their utility for better understanding of the traffic will also be discussed. The Transport Engineer can reuse them or apply TAU for extraction of other types of \textcolor{revision_1}{insights}. The source code used to extract these 17 types of \textcolor{revision_1}{insights} is shared with the community\footnote{The source code is planned to be shared after the publication of the paper.}.

\section{Experimental results}\label{section:experimental-results}
In this section, TAU is applied to extract 17 other types of \textcolor{revision_1}{insights}. They are enumerated in  Table \ref{tab:other_types_of_knowledge}. The enumeration starts with the number \(8\). In fact, the first 7 are the main types of \textcolor{revision_1}{insights} in TAU and they were described above.    
\par
\begin{table}[!h]
\centering 
\begin{tabular}{ll}
\hline
\textbf{N} & \textbf{\textcolor{revision_1}{Insights'} type}                                                                                                            \\ \hline
\textbf{8}          & \begin{tabular}[c]{@{}l@{}}Number of Vehicles that crossed every line apart from \\ both directions.\end{tabular}                   \\ \hline
\textbf{9}         & \begin{tabular}[c]{@{}l@{}}The vehicle trajectories followed by all the vehicles \\ during the recorded UAV video.\end{tabular}    \\ \hline
\textbf{10}         & \begin{tabular}[c]{@{}l@{}}The total number of vehicles in all the recorded area \\ over time.\end{tabular}                        \\ \hline
\textbf{11}         & The number of vehicles per zone over time.                                                                                         \\ \hline
\textbf{12}         & \begin{tabular}[c]{@{}l@{}}Average velocity of vehicles in all the recorded area\\ over time.\end{tabular}                         \\ \hline
\textbf{13}         & Average velocity of vehicles per zone over time.                                                                                   \\ \hline
\textbf{14}         & \begin{tabular}[c]{@{}l@{}}Number of Allowed and Forbidden crosses over time\\ during the recorded session.\end{tabular}           \\ \hline
\textbf{15}         & \begin{tabular}[c]{@{}l@{}}Correlation degree between the Average velocity and\\ the number of vehicles over time.\end{tabular}    \\ \hline
\textbf{16}         & \begin{tabular}[c]{@{}l@{}}Factors influencing the number of allowed and\\ forbidden crosses over time\end{tabular}                \\ \hline
\textbf{17}         & Heat map of the maximum recorded velocity per pixel                                                                                \\ \hline
\textbf{18}         & Heat map of the congestion level per pixel                                                                                         \\ \hline
\textbf{19}         & Heat map of the average velocity of vehicle per pixel.                                                                             \\ \hline
\textbf{20}         & \begin{tabular}[c]{@{}l@{}}The diagram of vehicle movements between zones \\ during the recorded UAV video. \end{tabular}                      \\ \hline

\textbf{21}         & \begin{tabular}[c]{@{}l@{}}The \textcolor{eaai_rev_1}{pdf} (Probability Density Function) of vehicles \\ sizes during the recorded UAV video.\end{tabular} \\ \hline
\textbf{22}         & \begin{tabular}[c]{@{}l@{}}The \textcolor{eaai_rev_1}{pdf} (Probability Density Function) of vehicles \\ velocities during the UAV video.\end{tabular}     \\ \hline
\textbf{23}         & \begin{tabular}[c]{@{}l@{}}The histogram of vehicle sizes distribution per frame.\end{tabular}     \\ \hline
\textbf{24}         & \begin{tabular}[c]{@{}l@{}}The histogram of vehicle velocities distribution \\ per frame.\end{tabular}     \\ \hline
\end{tabular}
\caption{\small \sl Description of 17 of other types of \textcolor{revision_1}{insights} that can be extracted from TAU.
\label{tab:other_types_of_knowledge}} 
\end{table}
The types of \textcolor{revision_1}{insights} extracted from TAU are not limited to the types described in Table \ref{tab:other_types_of_knowledge}. Much more insights can be extracted following the needs of the Transport Engineer. \par
\textcolor{eaai_rev_1}{In the following subsections, we will first measure the accuracy of the vehicle detection task in TAU. This task is the base for all insights concluded later. Then, we will show an example of extraction of the seven main types of insights in TAU. After that, we will show, one by one,  the result of extraction of the other 17 types of insights described in Table \ref{tab:other_types_of_knowledge}. }

\subsection{\textcolor{eaai_rev_1}{Vehicle Detection Accuracy in TAU}}
\textcolor{eaai_rev_1}{To estimate the correctness of the TAU statistics, we should first assess the accuracy of the vehicle detection task. Thus, we constructed an object detection dataset and labeled five types of classes (car, bicycle/motorcycle, bus, truck and pedestrian). Later, the class pedestrian is kept for future studies and the bounding boxes related to this class are omitted from the final visualizations. After the dataset construction, we trained YOLO v3 \cite{YOLOv3} on it. } \par
\textcolor{eaai_rev_1}{To measure the Vehicle detection accuracy, we used the precision and recall metrics, which are defined as follows:
\begin{equation} \label{eq:metrics1}
recall = \frac{TP}{TP+FN}
\end{equation}
\begin{equation} \label{eq:metrics2}
precision = \frac{TP}{TP+FP}
\end{equation}
where \(TP\) is the true positives (correctly detected vehicles), \(FP\) is the false positives (false detection of vehicle) and \(FN\) is the false negatives ( vehicles not detected). 
} \par
\textcolor{eaai_rev_1}{Another metric is used to estimate the global performance of the vehicle detection task. This metric is the \(F1\) metric, which is a combination between the recall and precision metrics and defined in the following equation:
\begin{equation} \label{eq:metrics3}
F1 = \frac{2 \times recall \times precision}{recall + precision}
\end{equation}}
\textcolor{eaai_rev_1}{We also used the mean average precision metric (mAP). This metric was specifically designed for the object detection task. This metric estimates the global performance of a model under multiple confidence thresholds. To calculate this metric, we pass by two stages. The first stage is the calculation of the Average Precision for every separate class. AP is defined by the the following equation:
\begin{equation} \label{eq:metrics4}
AP = \sum_n (r_{n+1} - r_n)p_{interp}(r_{n+1})
\end{equation}
where 
\begin{equation} \label{eq:metrics5}
p_{interp}(r_{n+1}) = \max_{\tilde{r}:\tilde{r} \ge r_{n+1} p(\tilde{r})}
\end{equation}
with \(p(\tilde{r})\) is the precision calculated at the recall \(\tilde{r}\).
}
\textcolor{eaai_rev_1}{At the second stage, the mAP is concluded as the average of the separate APs calculated for every class. The formula is defined by the following equation:
\begin{equation} \label{eq:metrics6}
mAP = \frac{1}{N} \sum_{i=1}^N AP_i
\end{equation}
Where N is the number of classes. In real case, we don't use the full range of confidence thresholds. For example, we use mAP(0.5) which is the mAP calculated for the confidence threshold of 0.5. Also, we use mAP(0.5:0.95) which is the mAP calculated for different confidence thresholds starting from 0.5 to 0.95, by a step of 0.05. 
}
\textcolor{eaai_rev_1}{The figure \ref{fig:mAP_curve} draws the progress of the metric mAP (0.5) during the training of YOLO v3 \cite{YOLOv3} on the constructed vehicle detection dataset. 
\begin{figure*}[!h]  
\begin{center}  
\includegraphics[width=18cm]{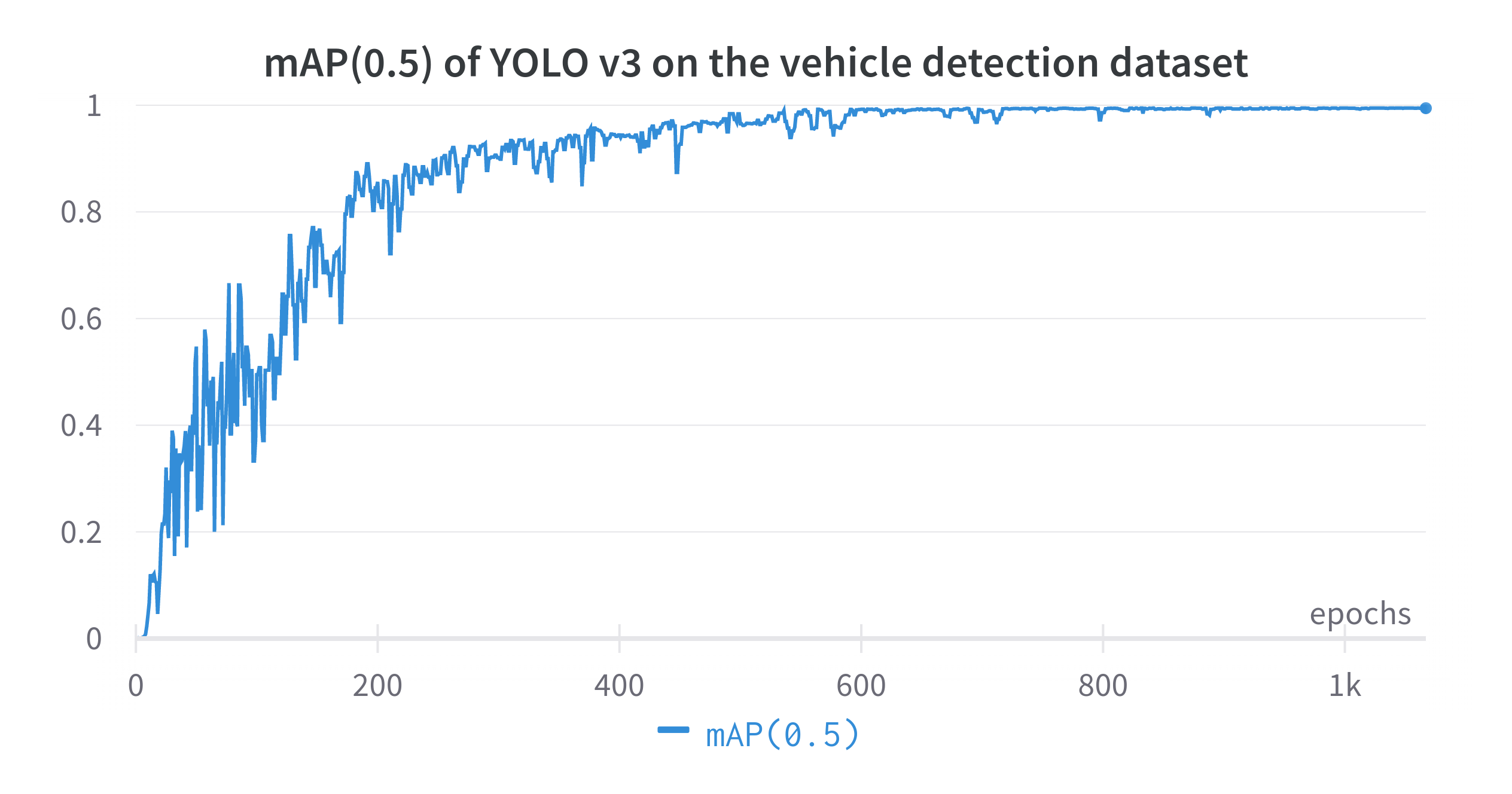}
\caption{\textcolor{eaai_rev_1}{The progress of the mAP (0.5) metric during the training of YOLO v3 on the constructed vehicle detection dataset.}  
\label{fig:mAP_curve}}  
\end{center}  
\end{figure*}}\par  
\textcolor{eaai_rev_1}{In table \ref{tab:yolov3_metrics}, we put the performance metrics of the vehicle detection task in TAU. We put the statistics for both YOLO v3 and YOLO v3 tiny (which is a small version of the model designed for embedded and real time devices). 
\begin{table}[!h]
\begin{center} 
\begin{tabular}{lll}
\hline
\textbf{}              & \textcolor{eaai_rev_1}{Yolo v3} & \textcolor{eaai_rev_1}{Yolo v3 tiny} \\ \hline
\textcolor{eaai_rev_1}{mAP(0.5)}      & \textcolor{eaai_rev_1}{0.9948}  & \textcolor{eaai_rev_1}{0.7767}       \\ \hline
\textcolor{eaai_rev_1}{mAP(0.5:0.95)} & \textcolor{eaai_rev_1}{0.8277}  & \textcolor{eaai_rev_1}{0.4451}       \\ \hline
\textcolor{eaai_rev_1}{Recall}        & \textcolor{eaai_rev_1}{0.9965}  & \textcolor{eaai_rev_1}{0.8045}       \\ \hline
\textcolor{eaai_rev_1}{Precision}     & \textcolor{eaai_rev_1}{0.9914}  & \textcolor{eaai_rev_1}{0.8822}       \\ \hline
\textcolor{eaai_rev_1}{F1}     & \textcolor{eaai_rev_1}{0.9939}  & \textcolor{eaai_rev_1}{0.8415}       \\ \hline
\end{tabular}
\caption{\textcolor{eaai_rev_1}{Performance metrics of the TAU vehicle detection task.}  
\label{tab:yolov3_metrics}}  
\end{center}
\end{table}
}\par
\textcolor{eaai_rev_1}{The metrics show a very high accuracy of the vehicle detection. This is mainly because the viewpoint from which we see the vehicles is the same for the whole dataset. This helps the model to grasp the pattern more accurately. Due to this high accuracy, all the insights based on it will have a similar accuracy. For example, the vehicle tracking task will be based on the bounding boxes estimated for the vehicle. There will not be identity switches because all the vehicles are moving on the same ground. Hence, we will have a good accuracy for the tracking task. Second, for vehicle assignment to zones, all the assignments will be based on the center of the bounding box of the vehicles. The accuracy of the zone assignment will be directly derived from the vehicle detection task. Thus, we will get a similar accuracy for it. This is similar for the other extracted insights.}
\subsection{Main \textcolor{revision_1}{insights} extracted from the UAV video.}
\begin{figure*}[!h]  
\begin{center}  
\includegraphics[width=18cm]{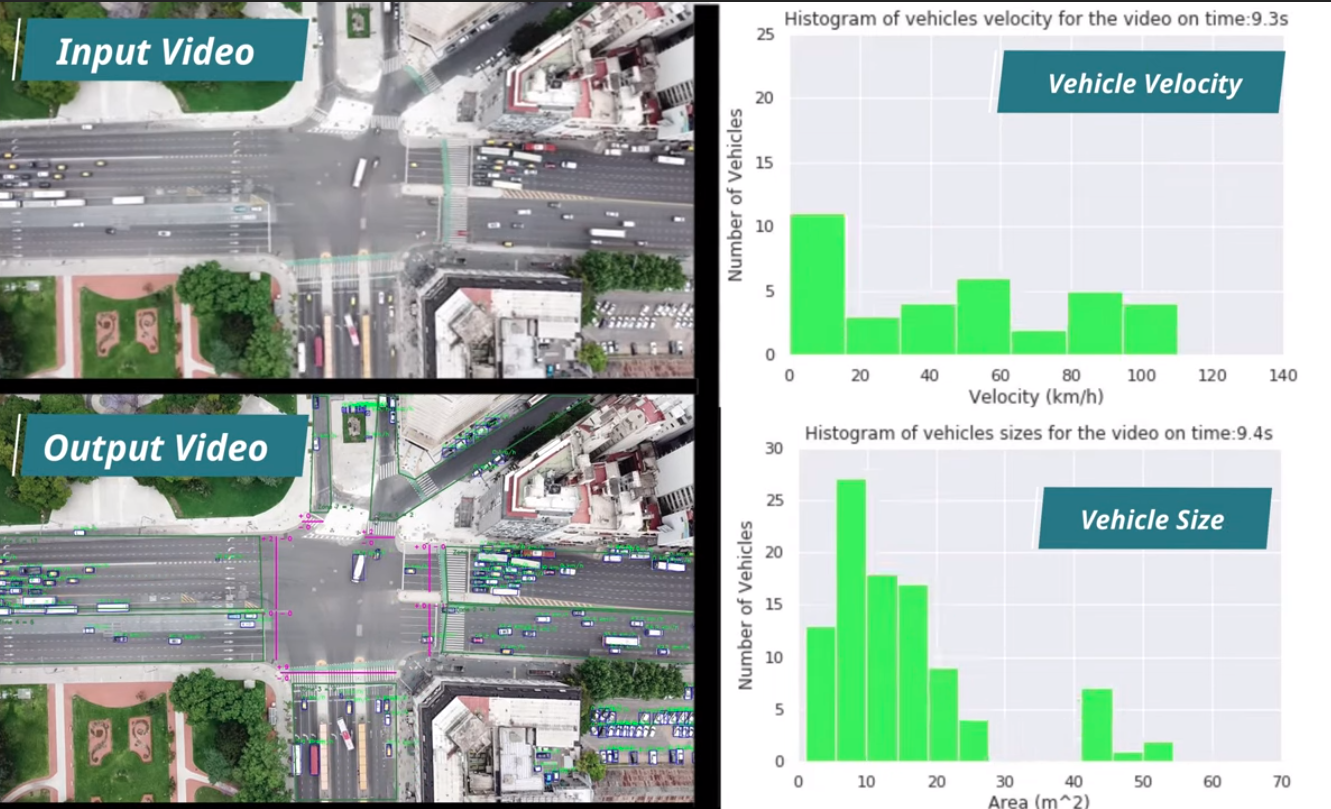}
\caption{\small \sl Example of \textcolor{revision_1}{insights'} extraction from an UAV video using TAU.  
\label{fig:tau_general_view}}  
\end{center}  
\end{figure*} 

The principle of TAU was elementary: providing a robust and straightforward pipeline to automatically analyze traffic using UAV video. In the Figure \ref{fig:tau_general_view} (the original video is presented here: https://youtu.be/kGv0gmtVEbI), we show an example of extracting two other  \textcolor{revision_1}{insights} from one UAV video. The figure represents four videos. The video on the top left corner represents the input UAV video taken on top of a crossroad. The video in the bottom left corner represents the results after applying the TAU framework to extract the seven main types of \textcolor{revision_1}{insights}. \textcolor{revision_1}{The two plots on the right} represent the extraction of two other types of \textcolor{revision_1}{insights}: the histogram of vehicle velocity per frame and the histogram of vehicle size per frame. \par
To get \textcolor{revision_1}{better visualization} on the output video generated by TAU, please see the following link: https://youtu.be/wXJV0H7LviU. For the remainder of the paper, this output video will be referenced as \emph{UAV test video}. A screenshot of the video is shown in Figure \ref{fig:main_knowledge_types}. 

It is clear that TAU generates the bounding box for every vehicle and estimates the velocity in km/h. Therefore, seven zones are defined by the user in this video. The number of vehicles inside every zone is updated after every frame. 
\subsection{Number of Vehicles that crossed every line apart from  both directions.}
\begin{figure}[!h]
\begin{center}  
\includegraphics[width=4cm]{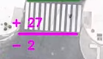}
\caption{\small \sl Number of Vehicles that crossed one line from both directions. 
\label{fig:car_crosses}}  
\end{center}  
\end{figure} 
If the Transport Engineer wants to study the vehicles' crosses over specific lines, he can implement it based on TAU. It will give him three pieces of valuable information. The first is the number of vehicles' crosses for the total UAV video. The second is the number of allowed crosses. The third is the forbidden crosses over this line. For example, Figure \ref{fig:car_crosses} shows that during 4:00 of the selected UAV video, a total of 29 crosses had been recorded: 27 are allowed, and two are forbidden. 

\subsection{The vehicle trajectories followed by all the vehicles during the recorded UAV video.}

\begin{figure}[!h]  
\begin{center}  
\includegraphics[width=10cm]{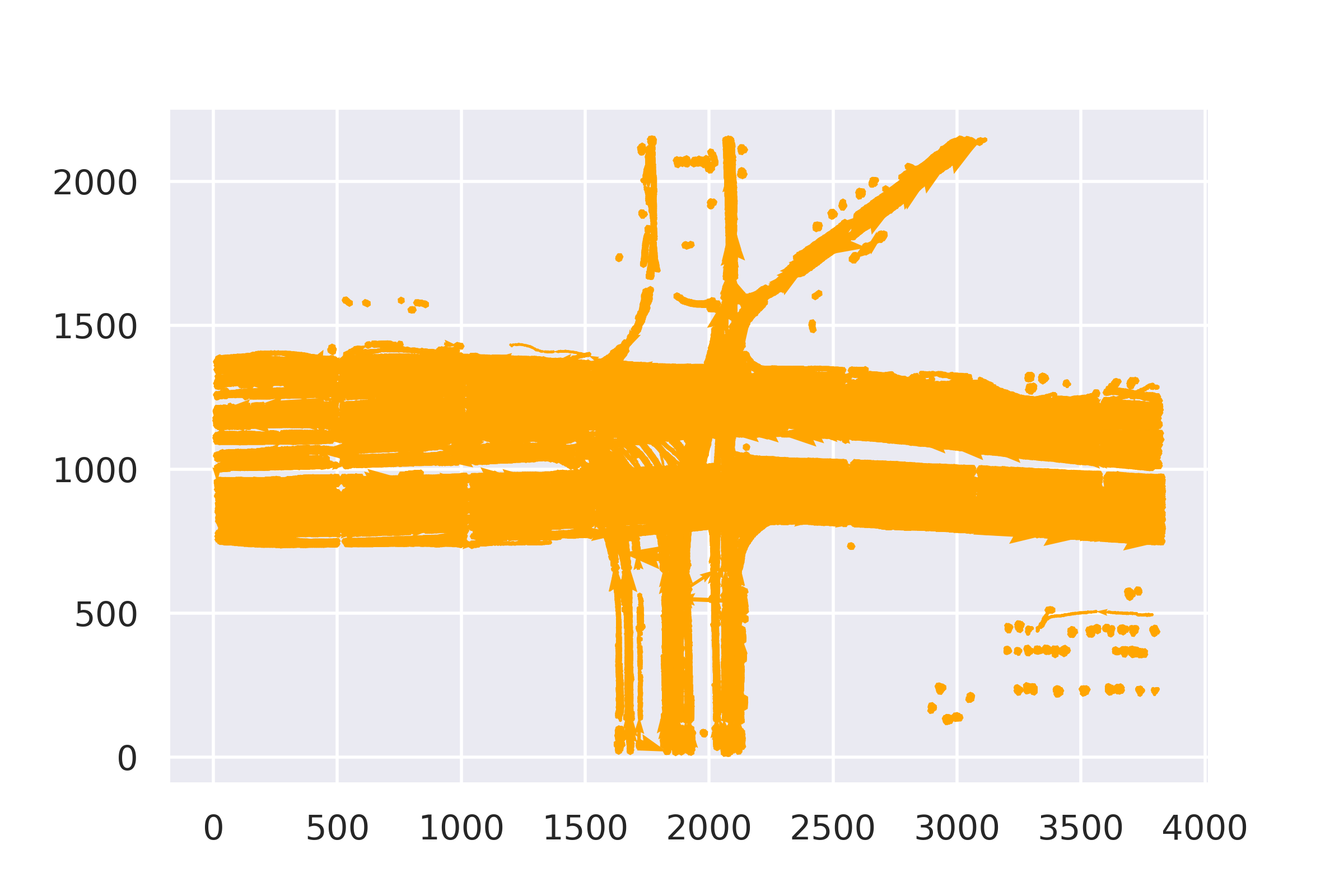}
\caption{\small \sl The total trajectories followed by all the vehicles that circulated inside the selected area during the recorded session. 
\label{fig:Trajectories}}  
\end{center}  
\end{figure} 
Based on TAU, we can draw all the trajectories followed by the vehicles inside a specific UAV video alongside their directions. For example, Figure \ref{fig:Trajectories} shows  these trajectories for the \emph{UAV test video}. It gives better insight into the vehicles' movements. 

\subsection{The total number of vehicles in all the recorded area over time.}
\begin{figure}[!h]  
\begin{center}  
\includegraphics[width=10cm]{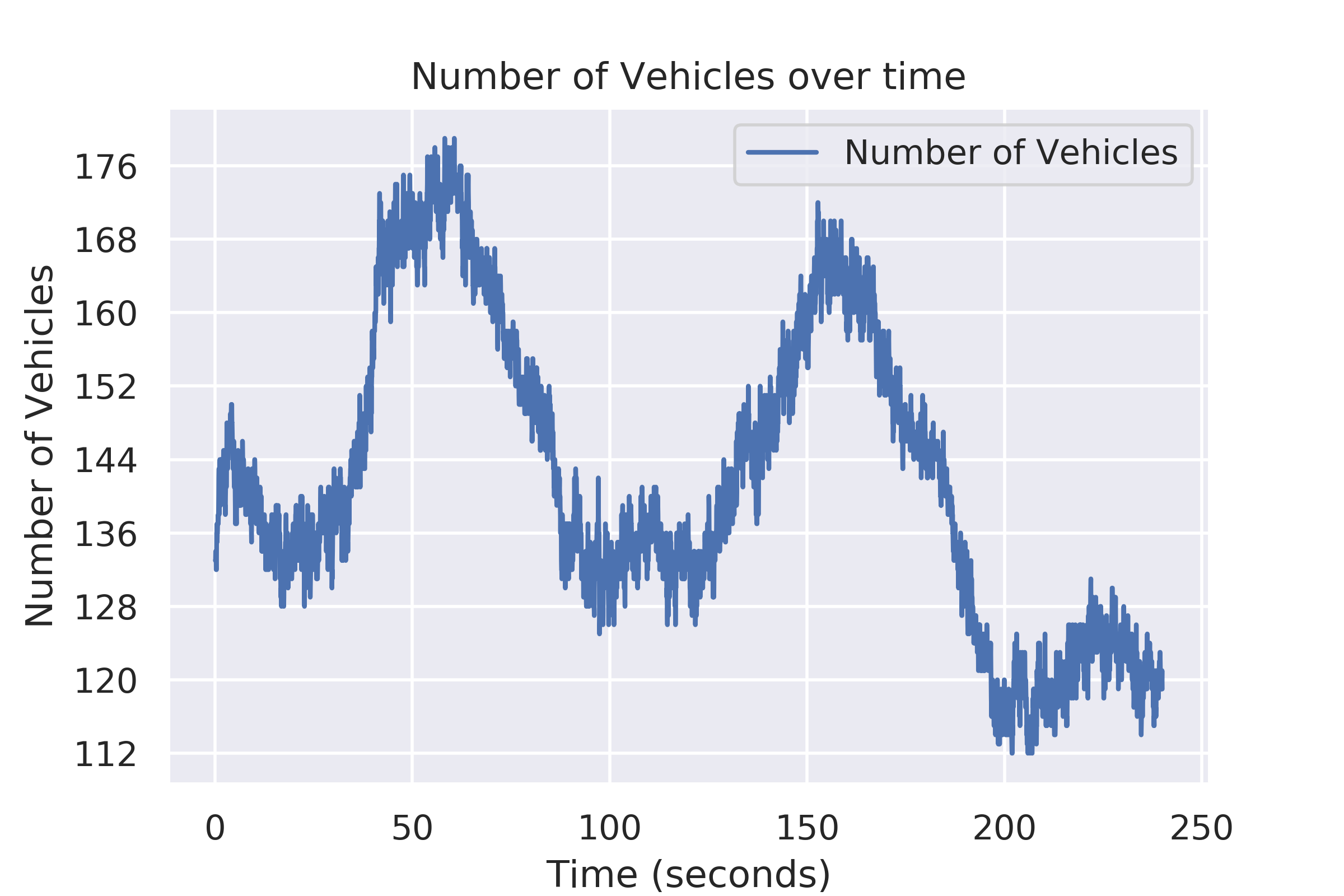}
\caption{\small \sl The number of vehicles inside the recorded area over time. 
\label{fig:Nb_vehicel_time}}  
\end{center}  
\end{figure} 
To know the level of traffic congestion, it is beneficial to know the number of circulating vehicles over time. For example, Figure \ref{fig:Nb_vehicel_time} illustrates the curve of the number of vehicles by time for the \emph{UAV test video}. The number of vehicles fluctuates between 112 and 176 in total. 

\subsection{The number of vehicles per zone over time.}
\begin{figure}[!h]  
\begin{center}  
\includegraphics[width=10cm]{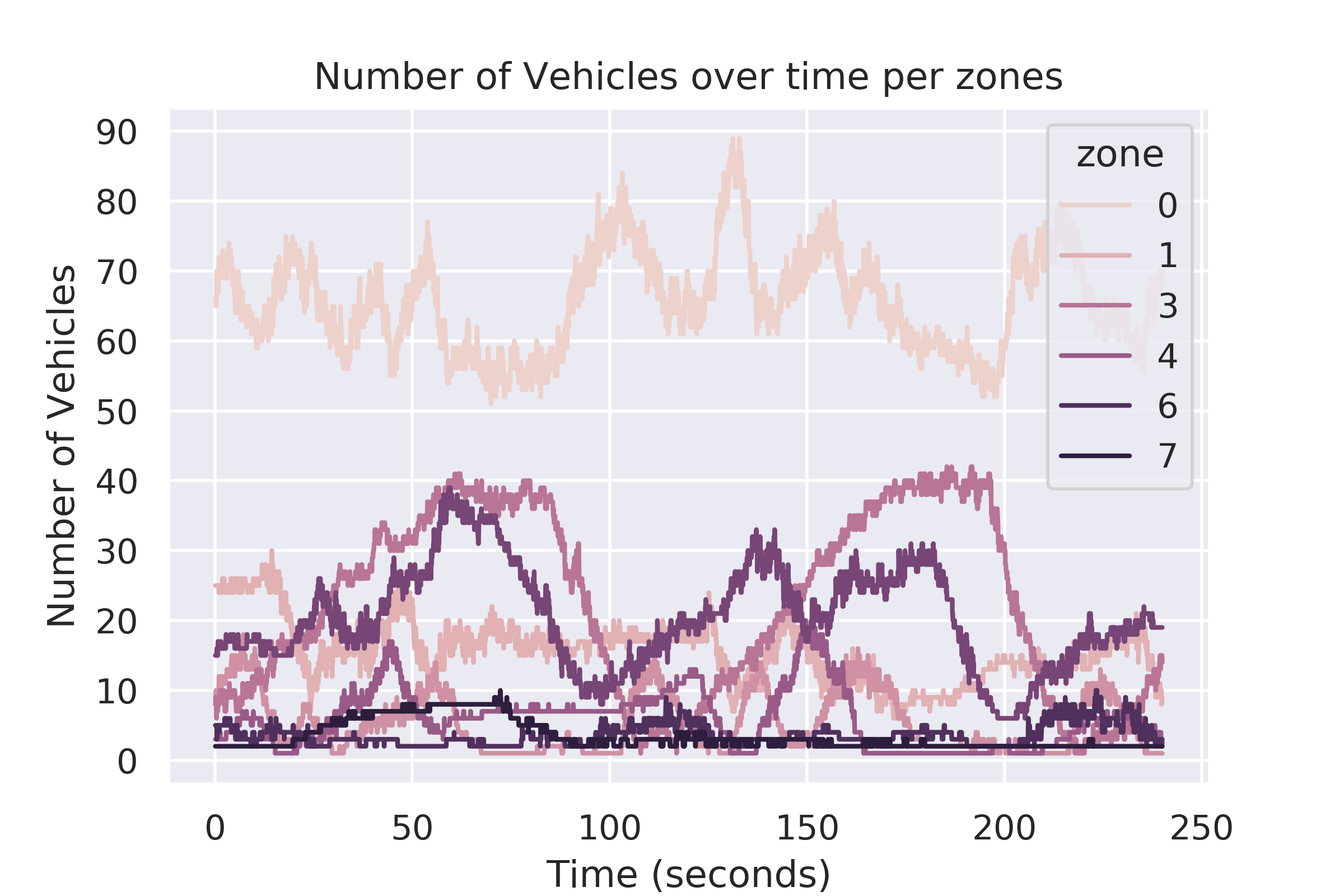}
\caption{\small \sl The number of vehicles inside the recorded area over time (per zones). 
\label{fig:Nb_vehicel_time_zone}}  
\end{center}  
\end{figure} 
Sometimes, it is advantageous to analyze the congestion level in specific zones and not in all the surveyed areas. This is to see if some zones are more crowded than others. For example, Figure \ref{fig:Nb_vehicel_time_zone} displays the curve of the number of vehicles for every zone by time for the \emph{UAV test video}. We can see that zones 6 and 7 are the less crowded while zones 3, 4, and 5 are the most crowded. Zone 0 means the areas outside the zones of interest defined by the user (zones 1 to 7). 
\subsection{Average velocity of vehicles in all the recorded area over time.}

\begin{figure}[!h]   
\begin{center}  
\includegraphics[width=10cm]{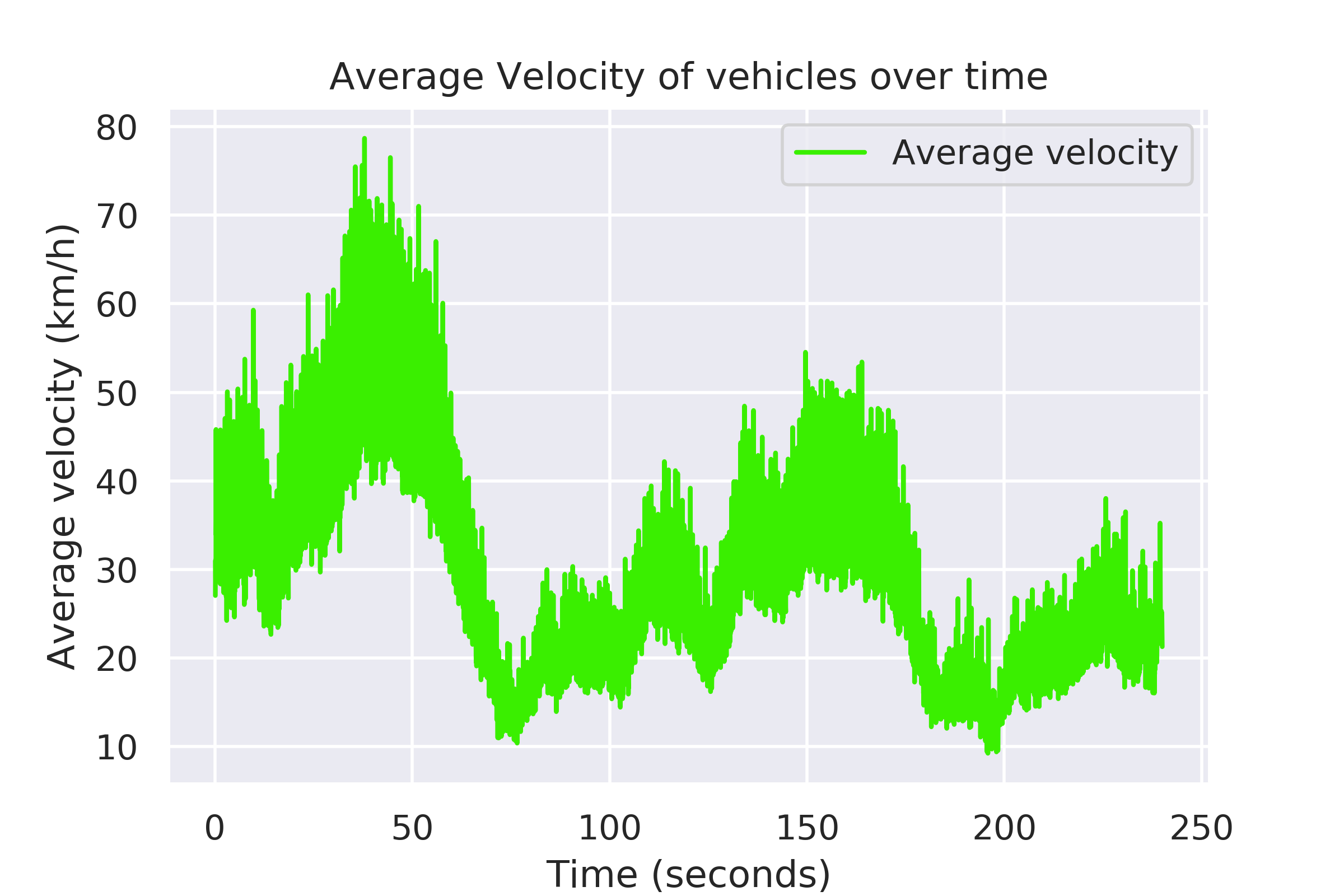}
\caption{\small \sl Average velocity of vehicles over time. 
\label{fig:av_vel_time}}  
\end{center}  
\end{figure} 
It is handy for the Transport Engineer to analyze the average velocity in all the surveyed areas. This metric reflects the degree of congestion and the quality of the traffic. If it is low, so there are some problems to resolve. If it is high but below the maximum allowed velocity, the traffic quality is good. If it is above the maximum allowed velocity, many drivers are violating the rules. For example, Figure \ref{fig:av_vel_time} illustrates the curve of the average velocity of all the vehicles circulating in the \emph{UAV test video} over time. We can see that the average velocity fluctuates between 10 km/h and 78 km/h. This differs following the traffic lights indication for every road and the number of vehicles circulating in every zone. This detail will be emphasized more in the following sub-section. 
\subsection{Average velocity of vehicles per zone over time.}

\begin{figure}[!h]   
\begin{center}  
\includegraphics[width=10cm]{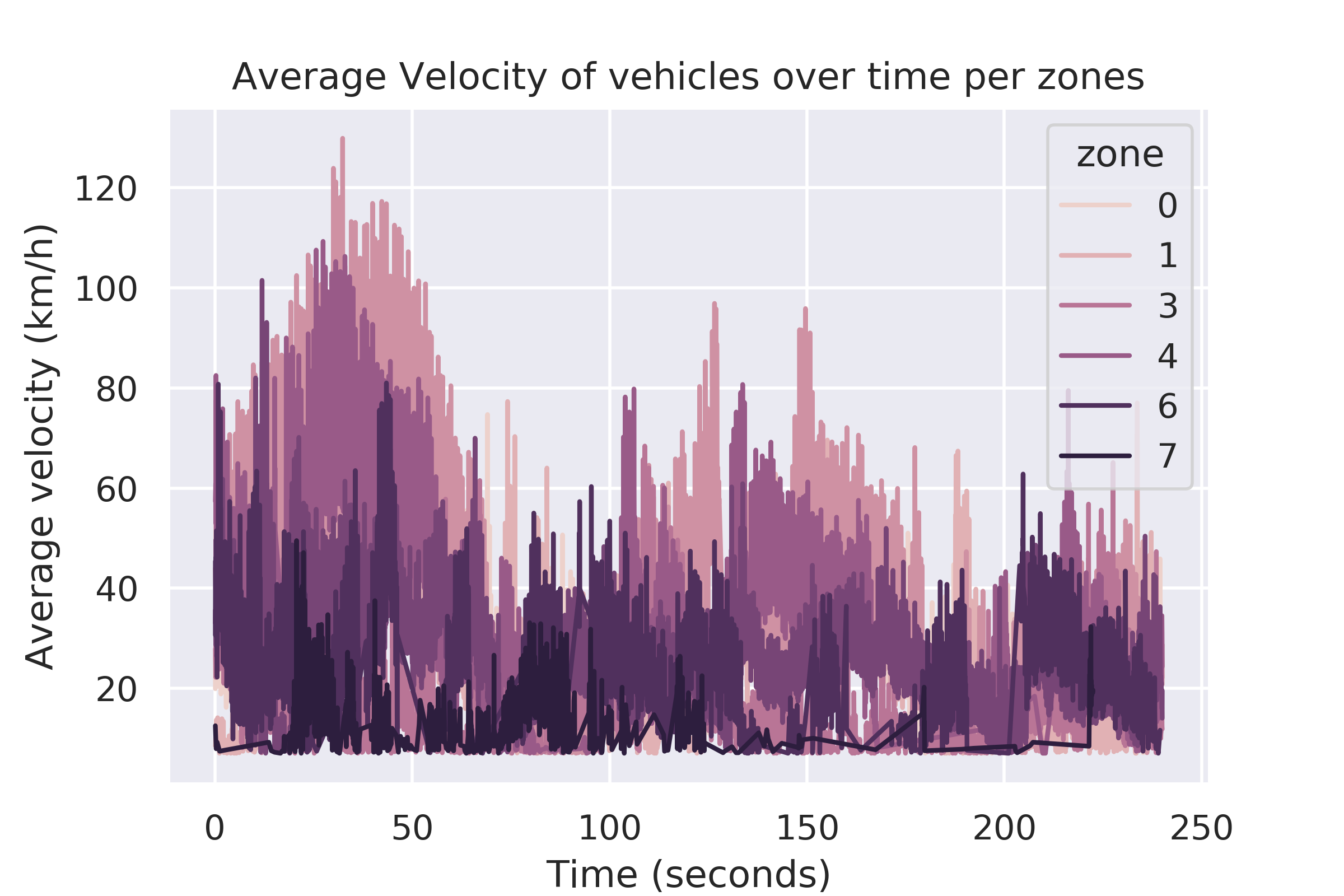}
\caption{\small \sl Average velocity of vehicles over time (per zones). 
\label{fig:av_vel_time_zones}}  
\end{center}  
\end{figure} 
Sometimes, it is not fair to analyze the overall average velocity as the traffic conditions may change from one zone to another. Hence,  it is better to analyze the average velocity for every zone apart. This helps to understand if some zones suffer from abnormal congestion that needs investigation. For example, we drew the average velocity per zone curve for the vehicles circulating in the \emph{UAV test video}. We can see that zone 1 has the best traffic flow. We can see also that vehicles are circulating at zones 6 and 7 with low velocity. We note that sometimes the average velocity returns to zero due to traffic condition (traffic light is red, for example, which oblige all the vehicles to stop). 
\subsection{Number of Allowed and Forbidden crosses over time during the recorded session.}
\begin{figure}[!h]   
\begin{center}  
\includegraphics[width=10cm]{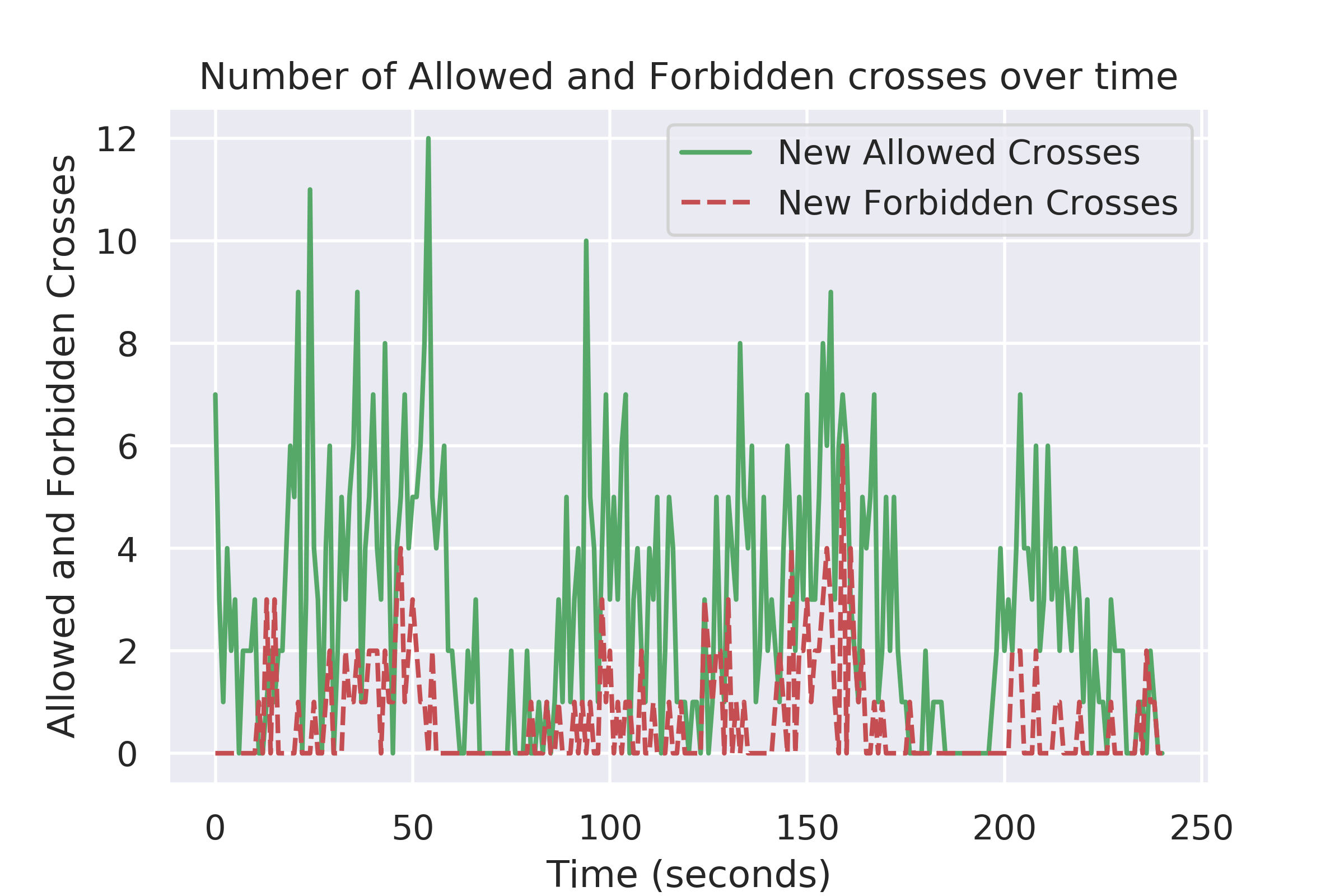}
\caption{\small \sl Number of Allowed and Forbidden crosses over time during the recorded session. 
\label{fig:Nb_Allowed_Forbidden_crosses}}  
\end{center}  
\end{figure} 
One of the most critical events in traffic is vehicles circulating in the wrong direction. This is dangerous driving behavior. Therefore, the Transport Engineer sometimes needs to go deeper in analyzing when this behavior occurs. This helps to understand the causes of such behavior and to resolve it. For example, Figure \ref{fig:Nb_Allowed_Forbidden_crosses} shows the curves of the new allowed and forbidden crosses for every second for the \emph{UAV test video}. We note, in general, that there is a correlation between the number of allowed and forbidden crosses. If the number of allowed crosses increases, the number of forbidden crosses also increases at the same pace. However, the total number of forbidden crosses increases unexpectedly between the seconds 140 and 160.  

\subsection{Correlation degree between the Average velocity and the number of vehicles over time.}
\begin{figure}[!h]   
\begin{center}  
\includegraphics[width=10cm]{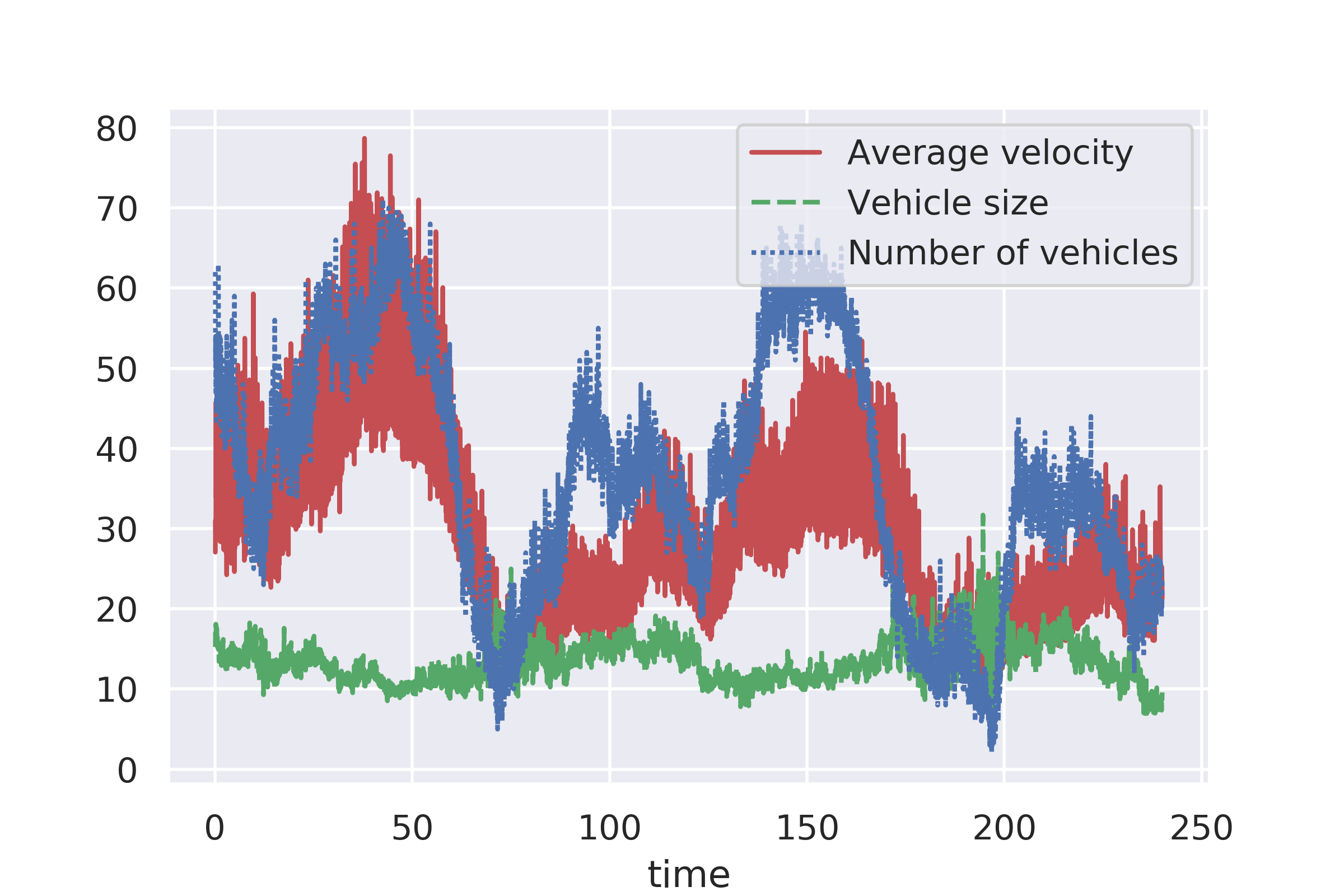}
\caption{\small \sl Correlation between the Average velocity and the Number of Vehicles over time. 
\label{fig:correlation_1}}  
\end{center}  
\end{figure} 
Sometimes,  the intuitive search for correlation between the different metrics is insufficient. Better is to use precise figures representing the different metrics and to use correlation metrics. For example, to interpret the relationship between the average velocity, the number of vehicles, and the vehicle sizes in the \emph{UAV test video}, we used two methods. First, we drew their representation in the same figure (please see the Figure \ref{fig:correlation_1}. Second, we calculated the Pearson correlation coefficient between them (Please see Table \ref{tab:pearson_correleation_1}).

\begin{table}[h] 
\centering
\resizebox{9cm}{!}{
\begin{tabular}{|l|l|l|l|}
\hline
                                                             & \begin{tabular}[c]{@{}l@{}}Average \\ Velocity\end{tabular} & \begin{tabular}[c]{@{}l@{}}Vehicle\\ size\end{tabular} & \begin{tabular}[c]{@{}l@{}}Number\\ of Vehicles\end{tabular} \\ \hline
\begin{tabular}[c]{@{}l@{}}Average \\ Velocity\end{tabular}  & \cellcolor[HTML]{036400}1.                            & -0.40                                              & \cellcolor[HTML]{009901}0.70                             \\ \hline
\begin{tabular}[c]{@{}l@{}}Vehicle\\ size\end{tabular}       & -0.40                                                   & \cellcolor[HTML]{036400}1.                       & -0.39                                                    \\ \hline
\begin{tabular}[c]{@{}l@{}}Number\\ of Vehicles\end{tabular} & \cellcolor[HTML]{009901}0.70                            & -0.39                                              & \cellcolor[HTML]{036400}1.                             \\ \hline
\end{tabular}}
\caption{\small \sl Pearson correlation coefficients between the values of Average Velocity, the Number of Vehicles and the the Vehicle sizes for the \emph{UAV test video}. 
\label{tab:pearson_correleation_1}} 
\end{table}
To remind the Pearson correlation coefficient, it is a real value that ranges between -1 and 1. The value 1 reflects a perfect positive correlation (like the correlation with the same factor). The value -1 reflects a perfect negative correlation. 0 means no correlation at all. Generally, the values from 0.50 to 1 mean a strong correlation. The values from 0.30 to 0.50 mean moderate correlation. The values between 0 and 0.30 mean low correlation. The meaning is also applicable in the negative scale for the degree of the negative correlation. From Table \ref{tab:pearson_correleation_1}, we can see that there is a high correlation between the Number of Vehicles circulating and the Average velocity (Pearson correlation coefficient = 0.70). This means that the more the average velocity increases, the more the number of vehicles are circulating and the more the quality of traffic flow improves. This correlation is consolidated in Figure \ref{fig:correlation_1}. On the other hand, there is a moderate correlation between these factors and the Vehicle sizes. Vehicle sizes do not interfere well with the quality of traffic flow in this case. Sometimes, some of these conclusions could be intuitive but proving them with measurable metrics is valuable. 

\subsection{Factors influencing the number of Allowed and Forbidden crosses over time.}

\begin{figure}[!h]   
\begin{center}  
\includegraphics[width=10cm]{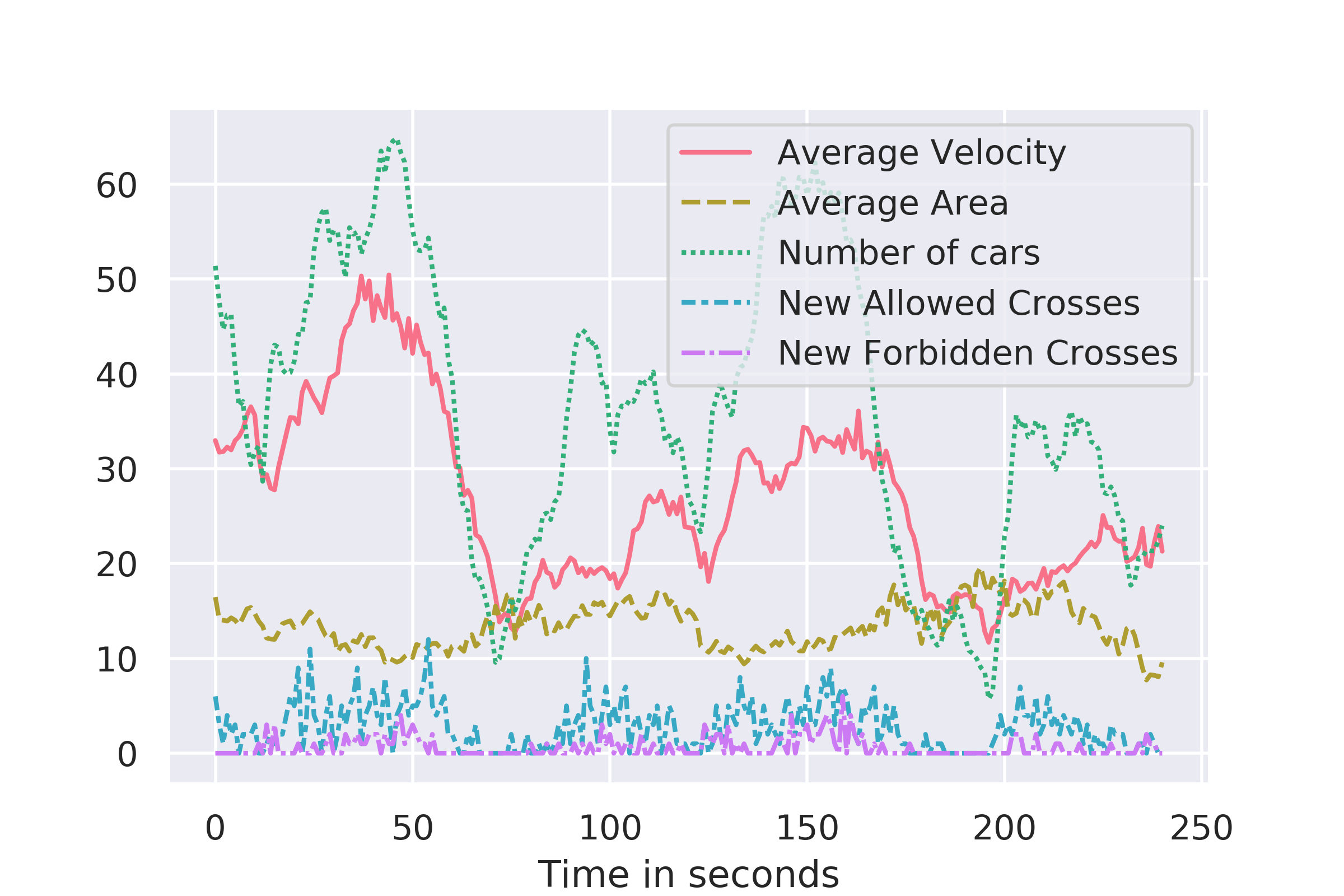}
\caption{\small \sl Correlation between different metrics related to Allowed and Forbidden crosses over time. 
\label{fig:correlation_3}}  
\end{center}  
\end{figure} 
To deeply analyze the factors leading to the increase of the forbidden crosses, the Transport Engineer can study the correlations between the new forbidden crosses on every second and the potential factors like the average velocity of vehicles, the average vehicle size, and the number of circulating vehicles. We applied this approach for the \emph{UAV test video} and drew the different curves of these factors in Figure \ref{fig:correlation_3}. The Pearson correlation coefficient between them is also calculated and displayed in Table \ref{tab:pearson_correleation_3}.

\begin{table}[]
\small
\resizebox{9cm}{4cm}{%
\begin{tabular}{|c|c|c|c|c|c|}
\hline
                                                                              & \textbf{\begin{tabular}[c]{@{}c@{}}Average \\ Velocity\end{tabular}} & \textbf{\begin{tabular}[c]{@{}c@{}}Average\\ size of\\ Vehicles\end{tabular}} & \textbf{\begin{tabular}[c]{@{}c@{}}Number \\ of \\ Vehicles\end{tabular}} & \textbf{\begin{tabular}[c]{@{}c@{}}New \\ Allowed \\ Crosses\end{tabular}} & \textbf{\begin{tabular}[c]{@{}c@{}}New \\ Forbidden \\ Crosses\end{tabular}} \\ \hline
\textbf{\begin{tabular}[c]{@{}c@{}}Average \\ Velocity\end{tabular}}          & 1.                                                                   & -0.50                                                                         & 0.78                                                                      & 0.44                                                                       & 0.30                                                                         \\ \hline
\textbf{\begin{tabular}[c]{@{}c@{}}Average\\ size of\\ Vehicles\end{tabular}} & -0.50                                                                & 1.                                                                            & -0.46                                                                     & -0.16                                                                      & -0.32                                                                        \\ \hline
\textbf{\begin{tabular}[c]{@{}c@{}}Number \\ of \\ Vehicles\end{tabular}}     & 0.78                                                                 & -0.46                                                                         & 1.                                                                        & 0.58                                                                       & 0.46                                                                         \\ \hline
\textbf{\begin{tabular}[c]{@{}c@{}}New \\ Allowed \\ Crosses\end{tabular}}    & 0.44                                                                 & -0.16                                                                         & 0.58                                                                      & 1.                                                                         & 0.34                                                                         \\ \hline
\textbf{\begin{tabular}[c]{@{}c@{}}New \\ Forbidden \\ Crosses\end{tabular}}  & 0.30                                                                 & -0.32                                                                         & 0.46                                                                      & 0.34                                                                       & 1.                                                                           \\ \hline
\end{tabular}}
\caption{\small \sl Pearson correlation coefficients between new forbidden crosses and a set of other factors for the \emph{UAV test video}. 
\label{tab:pearson_correleation_3}} 
\end{table}
Inside the Table \ref{tab:pearson_correleation_3}, we will only consider the column of "New Forbidden Crosses" and analyzes one by one the correlation with every potential factor. No strong correlation can be seen with them. Only a moderate correlation is noted. \par
First, we have a moderate correlation with the Average Velocity (0.30), the number of vehicles (0.46), and the number of Allowed Crosses (0.34). An increase in one of these factors has a medium influence on increasing the Number of Forbidden Crosses. The table shows that there is a strong correlation between these three factors. Hence, having more vehicles on the roads increases the chance to get forbidden crosses but with a moderate impact. \par
Second, A moderate negative relation is noted with the Average Size of vehicles. This means that bigger vehicles tend to behave more reasonably than smaller ones. This is because the size of the vehicle has a moderate impact. For example, it reduces its tendency to go in the wrong direction of the roads.

\subsection{Heat map of the maximum recorded velocity per pixel.} 

\begin{figure}[!h]   
\begin{center}  
\includegraphics[width=10cm]{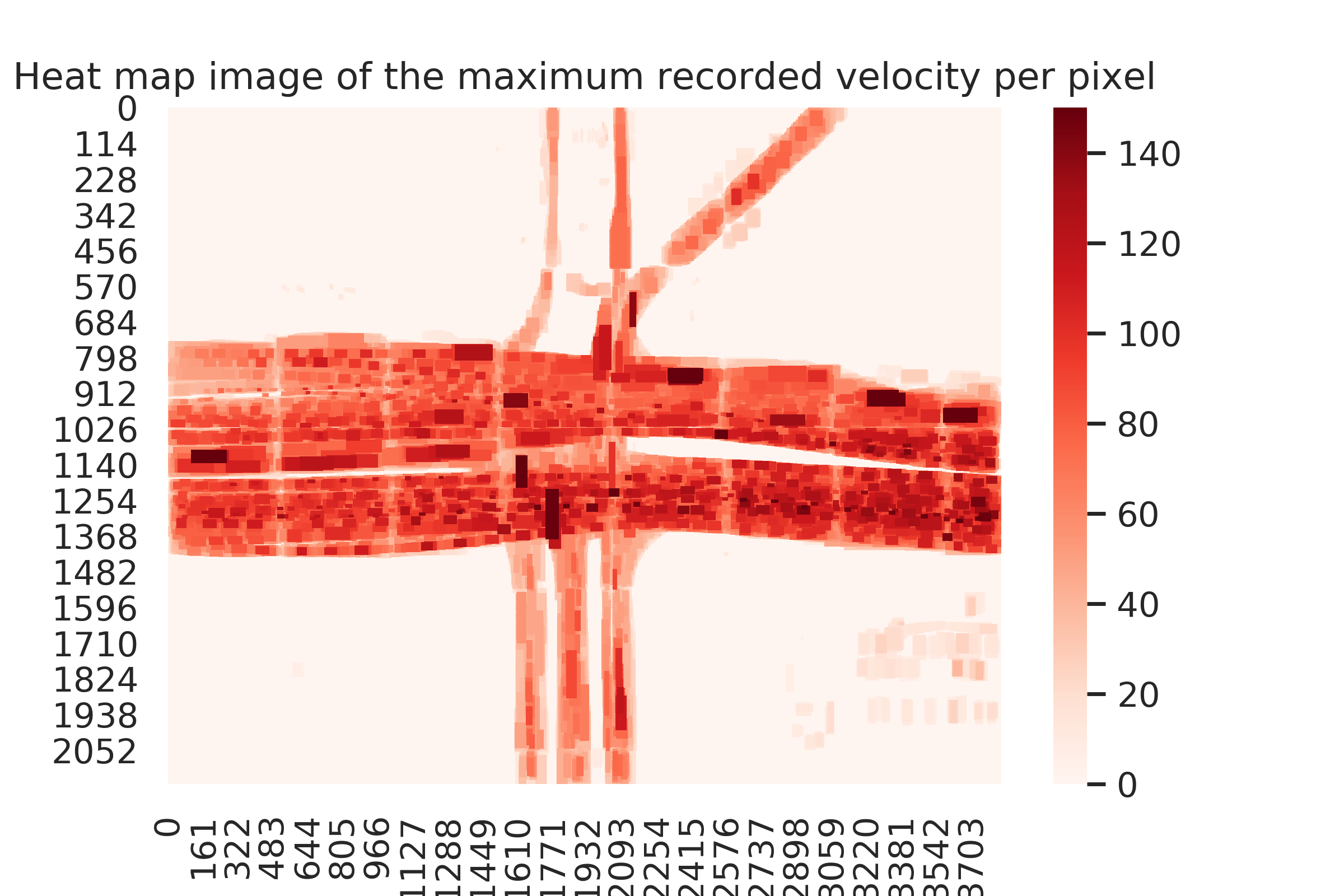}
\caption{\small \sl Heat map of the maximum recorded velocity per pixel during the recorded session. 
\label{fig:H_max_vel}}  
\end{center}  
\end{figure} 
Sometimes, having a large dataset about the vehicle movements does not give a tangible service to the Transport Engineer. He cannot know how to interpret them to understand the traffic. Therefore, it is recommended to use heat-maps. They give a beautiful visualization of the data to understand the different metrics recorded deeply. Hence, this study used three different heat maps to visualize the extracted data better. This subsection draws the maximum recorded velocity per pixel heat-map for all the surveyed areas. It is applied on the \emph{UAV test video} to get the Figure \ref{fig:H_max_vel}. \par
We can see in Figure \ref{fig:H_max_vel} that the value 0 is recorded for all the zones outside the roads. The non-null velocities are recorded only on the road's zone. It is also seen that the maximum recorded velocity varies from 20 km/h to 145 km/h. In some zones, the vehicles cannot surpass some velocity limits. Some vehicles reached very high velocities (more than 120 km/h). However, in general, the maximum velocity is an instant pic in the vehicle circulation and does not continue during all their trip. This is why we need to analyze the heat map of the average velocity to judge the vehicles' acceleration behavior. This is done in the following subsection. 

\subsection{Heat map of the average velocity of vehicles per pixel.}

\begin{figure}[!h]   
\begin{center}  
\includegraphics[width=10cm]{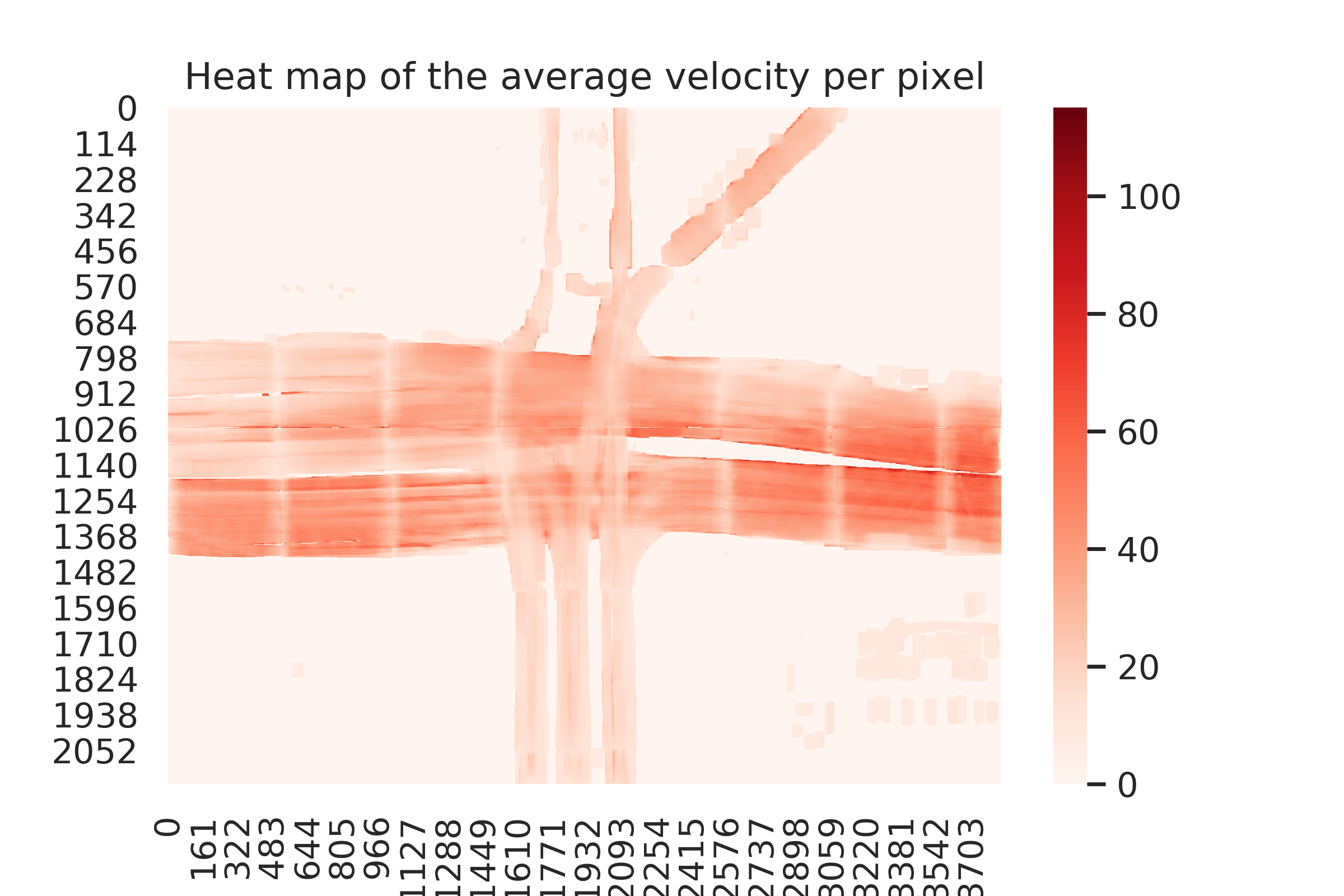}
\caption{\small \sl Heat map of the recorded average velocity per pixel during the session. 
\label{fig:H_avg_vel}}  
\end{center}  
\end{figure} 
The heat-map of the average velocity per pixel gives a precise measurement of the estimated velocity of one random vehicle circulating at every point on the roads. It is generated for the  \emph{UAV test video} to get the Figure \ref{fig:H_avg_vel}. It is noted that in some zones, the vehicles' movement is remarkably slower than in others. Vehicles tend to move very fast (above 60 km/h on average). The Transport Engineer could use this heat-map to ensure pedestrians' safety. Some zones can have high average velocity, while we may have pedestrians crossing it innocently. He can reduce the average velocity and protect these pedestrians by installing speed bumps in the critical parts. Sometimes, many Transport Engineers does not have the right mean to decide where to put the speed bumps. TAU gives them fast and precise metrics to decide better where to put them using only one UAV video.

\subsection{Heat map of the congestion level per pixel.}

\begin{figure}[!h]   
\begin{center}  
\includegraphics[width=10cm]{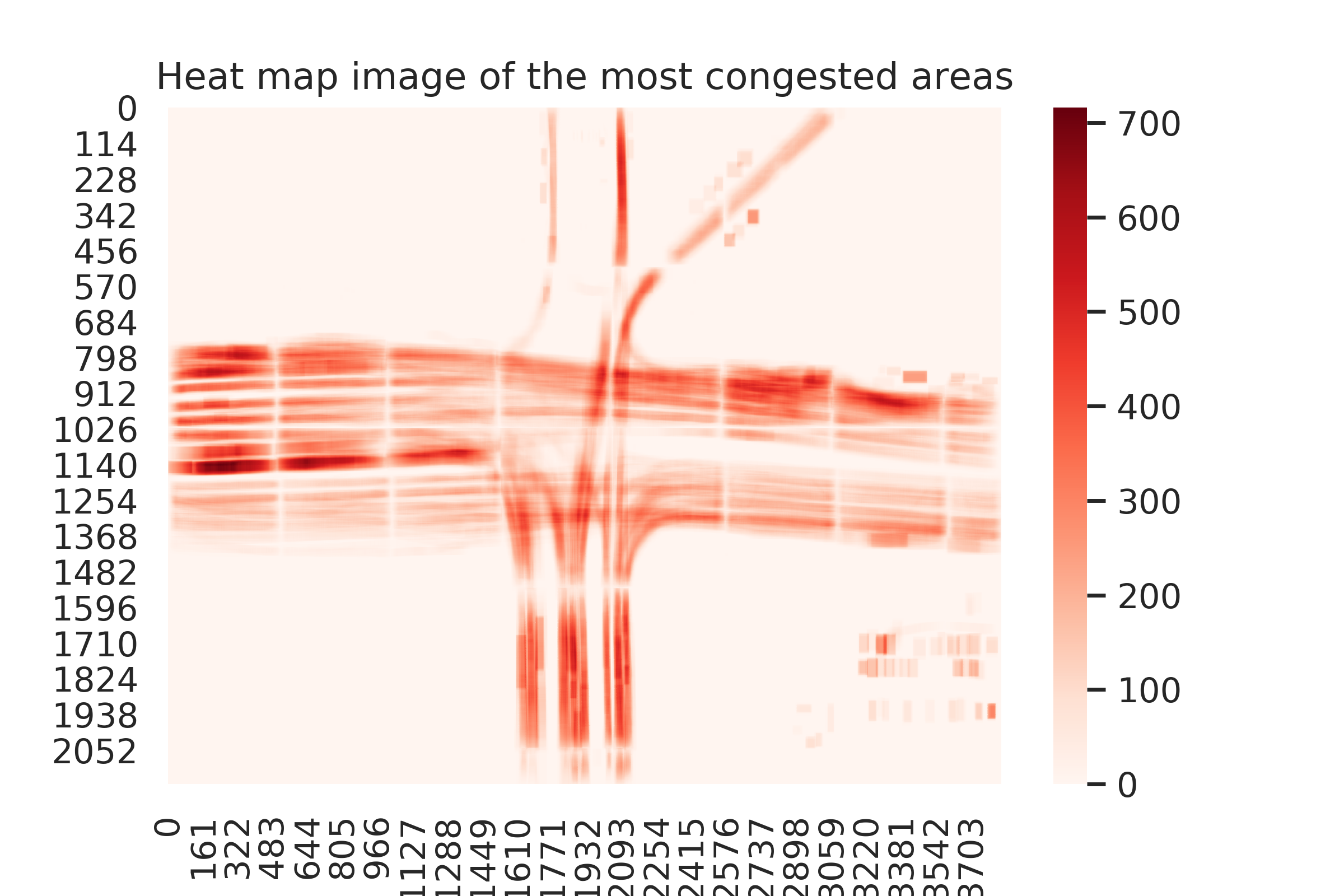}
\caption{\small \sl Heat map of road congestion level per pixel. 
\label{fig:H_cong_lev}}  
\end{center}  
\end{figure} 
Another useful heat-map that the Transport Engineer can use is the heat-map of the congestion level per pixel on the road. It reflects how many vehicles passed on every specific pixel of the surveyed area. It is applied for the  \emph{UAV test video} to get the Figure \ref{fig:H_cong_lev}. In most pixels, the value is 0, which means that no vehicles passed there. Also, most vehicles follow the same specific tracks on the roads. Most vehicles follow common tracks even for passing from one zone to another. This reflects that most vehicles respect the lines between tracks which is a good sign of driving awareness. Moreover, some specific tracks are very congested than others (more than 600 vehicles passed during the recorded video). Other tracks are rarely used (less than 50 vehicles passed during the recorded video). This gives clear insight into the congestion by focusing only on the congested tracks and trying to equilibrate the traffic load between all the tracks.

\subsection{The diagram of vehicle movements between zones during the recorded UAV video.}

\begin{figure*}  
\begin{center}  
\includegraphics[width=18cm]{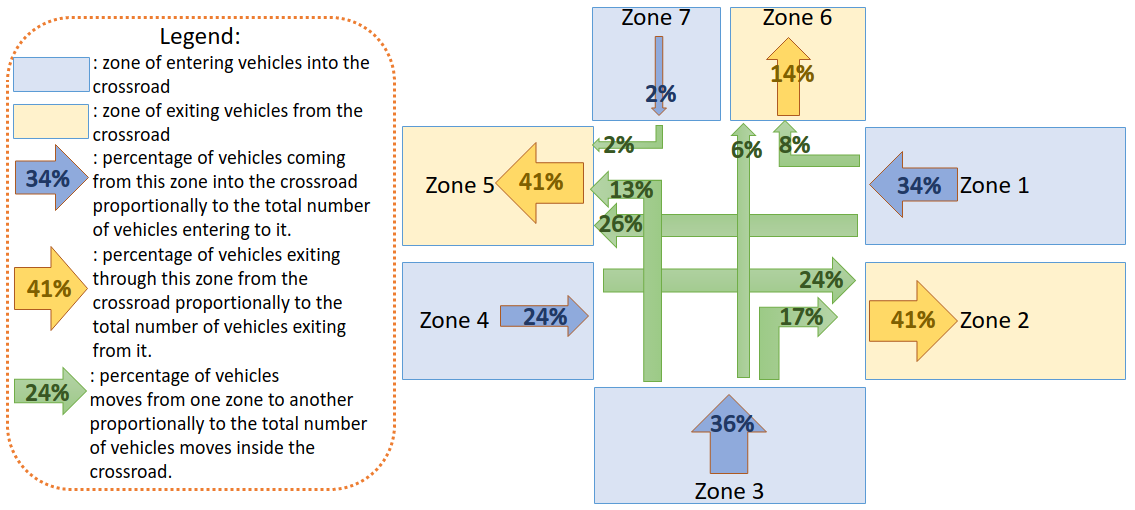}
\caption{\small \sl The Diagram of vehicle movements between zones during the recorded session. 
\label{fig:Diag_mvts_zones}}  
\end{center}  
\end{figure*} 
Crossroad management remains one of the biggest headaches for the Transport Engineer. Intersecting roads represent the most critical part of every traffic area. Every road can be divided into many tracks. Every track could have its direction. traffic lights are normally used to allocate time for every possible move inside the crossroad. Allocating time for the possible moves over the roads is a complicated process. A minimal error in this allocation could generate a significant waste of time for many drivers. TAU comes with a simple and efficient method to resolve all these problems. The method is based on counting the percentage of all the vehicles movements inside the crossroads. It is better to estimate it in a normal traffic use. This method gives a self-explainable metric to decide the right rules to optimize the traffic. To give an example, the method is run on \emph{UAV test video} to get the matrix represented in Table \ref{tab:matrix_crossroad_management}.

\begin{table}[h]
\resizebox{9cm}{!}{
\begin{tabular}{|l|l|l|l|l|l|l|l|}
\hline
\backslashbox{From}{To}   & \textbf{Z1} & \textbf{Z2}                & \textbf{Z3} & \textbf{Z4} & \textbf{Z5}                & \textbf{Z6}               & \textbf{Z7} \\ \hline
\textbf{Z1} & 0           & 0                          & 0           & 0           & \cellcolor[HTML]{FFCE93}26 & \cellcolor[HTML]{FFCE93}8 & 0           \\ \hline
\textbf{Z2} & 0           & 0                          & 0           & 0           & 0                          & 0                         & 0           \\ \hline
\textbf{Z3} & 0           & \cellcolor[HTML]{FFCE93}17 & 0           & 0           & \cellcolor[HTML]{FFCE93}13 & \cellcolor[HTML]{FFCE93}6 & 0           \\ \hline
\textbf{Z4} & 0           & \cellcolor[HTML]{FFCE93}24 & 0           & 0           & 0                          & 0                         & 0           \\ \hline
\textbf{Z5} & 0           & 0                          & 0           & 0           & 0                          & 0                         & 0           \\ \hline
\textbf{Z6} & 0           & 0                          & 0           & 0           & 0                          & 0                         & 0           \\ \hline
\textbf{Z7} & 0           & 0                          & 0           & 0           & \cellcolor[HTML]{FFCE93}2  & 0                         & 0           \\ \hline
\end{tabular}}
\caption{\small \sl Percentage of vehicle moves between the different zones in the crossroad represented in \emph{UAV test video}. 
\label{tab:matrix_crossroad_management}} 
\end{table}
The Table \ref{tab:matrix_crossroad_management} represents the percentage of the vehicle moves inside the crossroad represented in \emph{UAV test video}. The row index represents the starting zone of the vehicle. Zones are indexed from Z1 until Z7. Z0 is not represented because it is the default zone outside the user-defined zones. It does not have any specific meaning for crossroad management. The column index is the destination zone of the considered vehicle. For example, from the table: 17\% of the vehicles entering the crossroad pass from zone Z3 to zone Z2. The value is rounded to the nearest integer below the value. From this matrix, we drew a visually better diagram of the vehicles' moves in \(Figure\) \ref{fig:Diag_mvts_zones}. \par 
In \(Figure\) \ref{fig:Diag_mvts_zones}, the zones can be one of two types. The first type is the zones entering vehicles into the crossroad (Z1, Z3, Z4, and Z7). They are represented in blue color. Inside each of them, a blue arrow represents the percentage of vehicles entering from the selected zone into the crossroad. The second type of zones is the zones exiting the vehicles from the crossroad. They are represented in yellow color. Inside each of them, a yellow arrow represents the percentage of the vehicles exiting from the crossroad through the selected zone. The green arrows represent the percentage of the vehicle movements inside the crossroad. The thicknesses of drawn arrows follow the scales of the represented percentages. The diagram is drawn manually, but a software tool can be developed on top of TAU to generate it automatically from the UAV video like done for other types of \textcolor{eaai_rev_1}{insights (heatmaps, histograms, curves, and plots)}. \par
Using the Figure \ref{fig:Diag_mvts_zones}, the Transport Engineer can see in only one map all the traffic movements in the crossroad. Based on it, he can decide the different rules to set. For example, allocating time slots for the different traffic lights is not a headache like before. This allocation can be made following precisely the percentage of vehicle moves (Green Arrows). \par
TAU can be used for static crossroad management. Static means that the rules settings are not susceptible to continuous change. However, TAU is also useful for dynamic crossroad management. In dynamic crossroad management, the traffic rules can be changed continuously following the traffic conditions. For example, suppose the crossroad traffic is remarkably changed from one hour to another. In that case, the time allocation for the traffic lights can be changed following the new metrics recorded during that hour. A similar figure to Fig. \ref{fig:Diag_mvts_zones} will be generated for every hour to the transport engineer to change the rules based on it automatically. \par
For both static and dynamic crossroad management, TAU has given excellent service to analyze and decide the best rules to choose in every condition. Moreover, TAU has given self-explainable metrics that alleviate the responsibility from the side of the Transport Engineer because it is totally based on the recording of vehicles' moves on the roads. \par
\subsection{The probability density function of vehicles sizes during the recorded UAV video.} 

\begin{figure}[!h]   
\begin{center}  
\includegraphics[width=10cm]{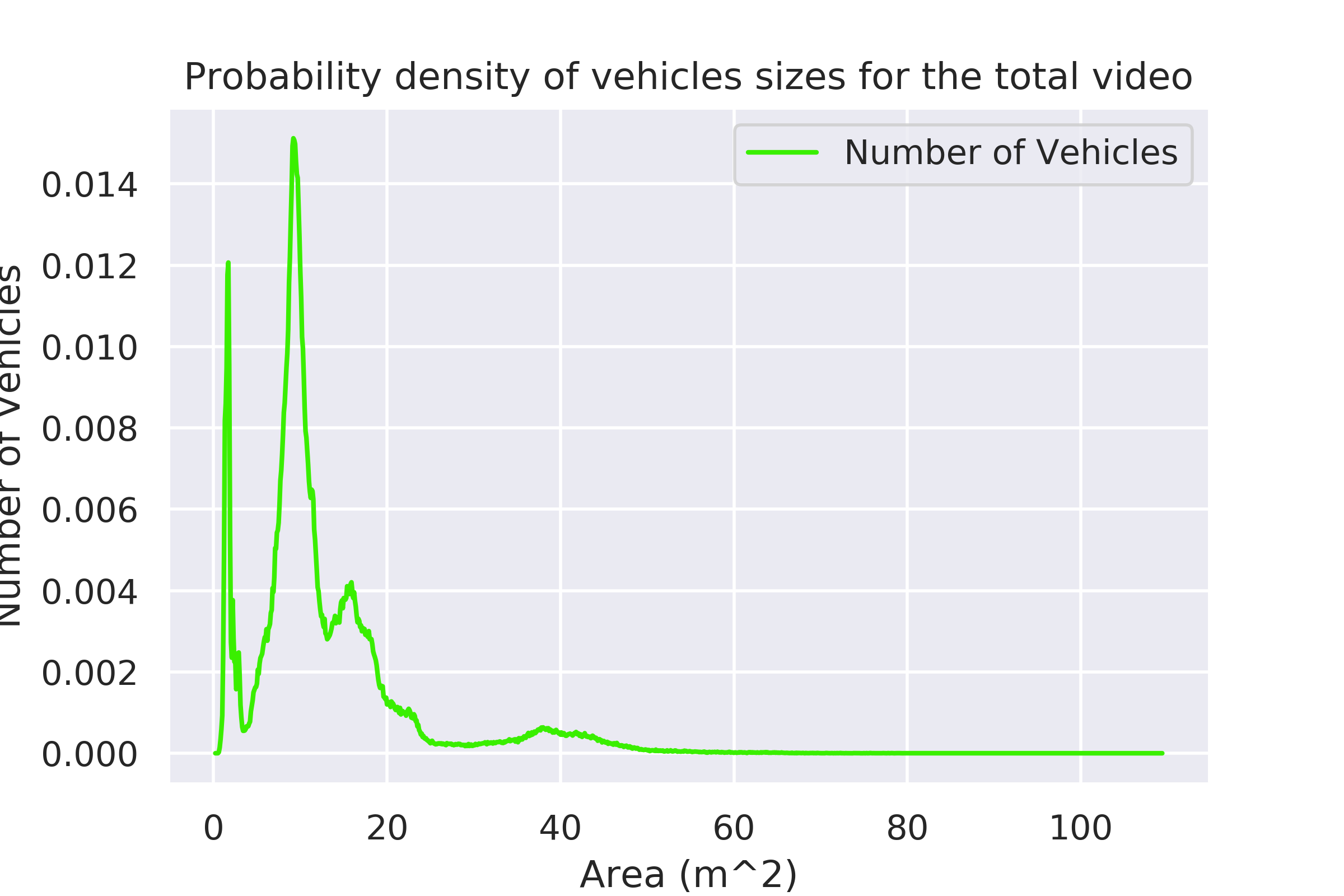}
\caption{\small \sl Probability density of vehicles sizes inside the selected area and the recorded session. 
\label{fig:Pr_Dens_Veh_size}}  
\end{center}  
\end{figure} 
It is valuable to know the distribution of vehicle sizes circulating in a surveyed area. This is why TAU has given the possibility to extract the Probability Density Function (\textcolor{eaai_rev_1}{pdf}) of the vehicle sizes from the UAV video. This feature is tested on the  \emph{UAV test video} to get the Figure \ref{fig:Pr_Dens_Veh_size}. Inside this figure, there are four extremums in the curve. The greatest extremum represents the most circulating vehicles in the video: sedan cars. The next three extremums are related to small-sized vehicles (like motorcycles), trucks, and buses. It is easy to generate the percentage of every size in the surveyed area from this \textcolor{eaai_rev_1}{pdf}. 

\subsection{Probability density function of vehicles velocities during the UAV video.}

\begin{figure}[!h]   
\begin{center}  
\includegraphics[width=10cm]{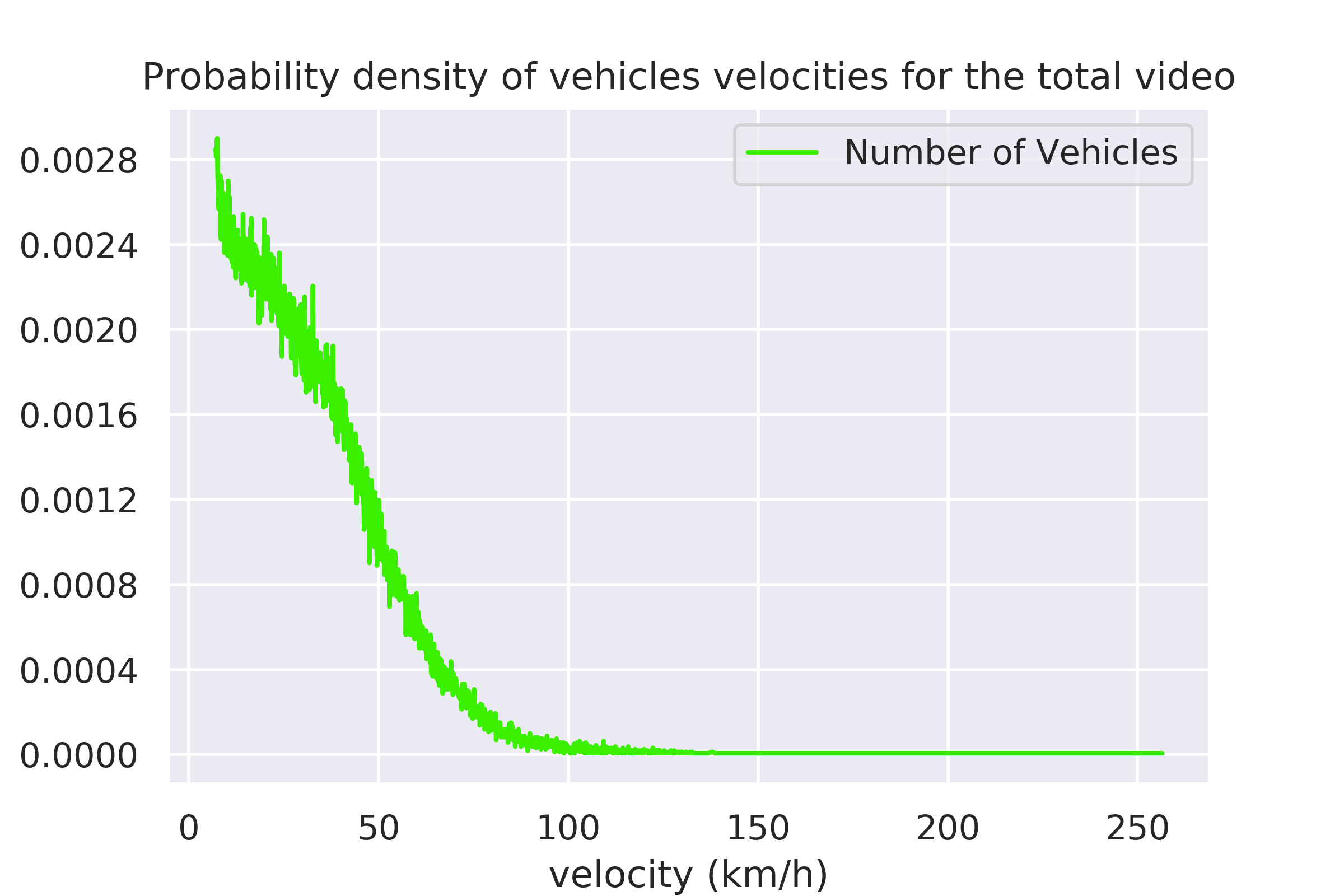}
\caption{\small \sl Probability density of vehicles velocity inside the selected area and the recorded session. 
\label{fig:Pr_Dens_Veh_vel}}  
\end{center}  
\end{figure} 
TAU has also given the possibility to generate the \textcolor{eaai_rev_1}{pdf} of vehicle velocities for one surveyed area. It gives the distribution of velocities recorded in the selected UAV video in one figure. This feature is tested on the \emph{UAV test video} to get the Figure \ref{fig:Pr_Dens_Veh_vel}. We can see that the most recorded velocities are below 50 km/m. There were minimally recorded velocities between 50 km/m and 90 km/m. Sporadic cases are recorded for velocities more than 90 km/h. This feature is handy for the Transport Engineer to reasonably set the speed limit on the roads without affecting the traffic's fluidity too much. 

\subsection{The \textcolor{eaai_rev_1}{histogram} of vehicle sizes distribution per frame}

Sometimes, getting one \textcolor{eaai_rev_1}{pdf} for the total video hides important changes occurring during the time. This is why TAU has given the possibility to generate \textcolor{eaai_rev_1}{histograms} for vehicle sizes distribution during the time. We found that representing the vehicle size distribution using a \textcolor{eaai_rev_1}{histogram} is more appropriate than \textcolor{eaai_rev_1}{{pdf}} to track changes over time easily. An example of this feature applied on the \emph{UAV test video} is presented in \(Figure\) \ref{fig:tau_general_view} (video: https://youtu.be/kGv0gmtVEbI). The histogram of the vehicle sizes is displayed in the lower right corner of the Figure. We can see that the global pattern of the histogram follows, in most cases, the full \textcolor{eaai_rev_1}{pdf} represented earlier. The same distribution of the vehicle size is kept, although there are continuous fluctuations. This can guide the Transport Engineer in setting the different sizes of vehicles using the traffic and the good rules to optimize it.

\subsection{The \textcolor{eaai_rev_1}{histogram} of vehicle velocities distribution per frame}

TAU  generates \textcolor{eaai_rev_1}{histograms} for the distribution of vehicle velocities over time. An example of this feature applied on the \emph{UAV test video} is presented in \(Figure\) \ref{fig:tau_general_view} (video: https://youtu.be/kGv0gmtVEbI). The histogram of the vehicle velocities is displayed on the right top corner of the Figure. This feature is handy for the Transport Engineer to study the different velocities recorded over time in the surveyed area. It helps estimate the number of vehicles breaking the speed limit and the number of vehicles respecting it. It can give a clear insight into the utility of setting new speed limits or removing old ones.

\subsection{\textcolor{revision_1}{Inference speed and hardware architecture}}
\textcolor{revision_1}{TAU is tested on a sample 4K video (resolution is 3840 × 2160). This is to measure the inference speed. The chosen sample video is taken from a fixed UAV  at 175 meters from the ground. The camera view is completely perpendicular to the ground. The input image is divided into 40 images of size \(512 * 512\) during the processing. Every image passes into the YOLO v3 network to catch the bounding boxes around every detected vehicle on the ground. Then we conclude the global parameters to be passed into the DeepSORT algorithm. During the generation of the output video, the inference speed of the YOLO v3 is always fixed. However, the inference speed of DeepSORT fluctuates following the number of vehicles. The more vehicles we have, the more we need processing power to calculate the Kalman predictions and the appearance descriptor for every vehicle. However, the overhead of the DeepSORT is far lesser than the one of YOLO v3. The machine that we used during the inference has the following specifications:
\begin{itemize}
  \item CPU: Intel(R) Core(TM) i7-7700HQ CPU @ 2.80GHz
  \item GPU: NVIDIA GeForce GTX 1060 6 GB VRAM
  \item Memory: 32 GB
  \item CUDA version: 9.1
  \item Framework: Tensorflow 1.14 (with Keras API)
\end{itemize}
We found that the average inference time needed to predict one frame is 4.3 seconds, corresponding to 0.23 FPS (Frames per Second). As the input video has 24 frames per second, processing one second of the UAV video costs one minute and 43 seconds. Therefore, we need  103 hours to process one hour from the UAV video. This high processing cost is due primarily to the YOLO v3 bottleneck. This bottleneck can be solved by working on larger input sizes,  parallelizing the processing, or considering fewer frames to process per second. 
}

\subsection{Limitations in the current version of TAU (TAU v1).}

During the test of the different features of TAU, we noted that there are some limitations to be targeted in the next version of the Framework (TAU v2):

\begin{itemize}
    \item Algorithms 1 \& 2 are not optimal: Although Algorithms 1 \& 2 did the job well, one drawback needs investigation. This drawback is the discontinuity of the metrics in pixels of coordinate \(x\) or \(y\) that are multiples of the size considered in processing. In the test, it was 512. This is the reason for the discontinuity in the metrics where vehicles pass by pixels multiple of 512. Both Algorithms 1 \& 2 need to be modified to overcome this flaw.
    \item TAU cannot run online in high resolutions: In the current version, TAU can be run online only for resolutions that are less than \(608*608\). The processing becomes a lot heavier with greater resolutions. Therefore,  parallelizations method are needed, especially for the heavier part of the processing (Algorithm 2). We need to divide the processing over multi-GPUs to get the different Vehicles Bounding Boxes for every processed frame. Then the Bounding boxes should be recalculated before running the Kalman filter and generating the appearance descriptors for every detected vehicle. Finally, the whole pipeline should be optimized to run on a multi-GPUs system to ensure the TAU framework's real-time running.
    \item TAU should take consideration for inclined vehicles: If the vehicle movement is following the \(\vec{x}\) axis or \(\vec{y}\), the bounding box is fitting the vehicle area allowing to adopt many approximations made in this study. However, if the vehicle moves following an inclined axis, the bounding box will not fit the vehicle, making the vehicle size calculation inappropriate. However, the velocity estimation remains efficient even in these conditions. 
\end{itemize}
\section{Conclusion}\label{section:conclusion}
This study introduced a robust and straightforward framework for traffic Analysis from UAV video (TAU). First, the scientific context of the study has been introduced. Then, the building blocks of the TAU have been described. After that, the theoretical basis of the seven main \textcolor{revision_1}{insights} extracted using the TAU from the UAV video have been detailed. It has been proven that the built Framework can be easily enriched to extract more \textcolor{revision_1}{insights}. In this study, 17 other \textcolor{revision_1}{insights} are taken as an example. In the experimental part, a total of 24 \textcolor{revision_1}{insights} from an UAV video sample have been discussed. We detailed the utility of these features for a transport engineer in his mission to optimize traffic fluidity, reduce waste of time and ensure safety on the roads. We tried to emphasize the main goal of TAU, which is to give the easiest way for a transport engineer to efficiently analyze and manage traffic by using only one UAV video, no more. \par
However, TAU has some limitations in its current version. The main limitation is the discontinuity of the metrics \textcolor{eaai_rev_1}{on the x-axes where the pixel index is multiple of the width of the processed frame, and also on the y-axes where the pixel index is multiple of the height of the processed frame.} The second main limitation is the inability to run online for high resolutions.\par

\textcolor{revision_1}{Beyond solving these limitations, TAU can be enriched by the addition of new features. For example, we can incorporate the predicted trajectory for every vehicle and estimate the recommended velocity for every vehicle in real-time. Also, architectural improvements can be made in TAU. For example, the chosen vehicle detector YOLO v3 \citep{YOLOv3} can be substituted by a more recent generic object detector like YOLO v4 \citep{bochkovskiy2020yolov4} or a designed object detector for aerial images specifically like the Butterfly detector \citep{adaimi2020perceiving}. Also, the vehicle tracker Deep SORT can be replaced with a more recent generic multi-object tracker (MOT) \citep{ciaparrone2020deep}. }\par
\textcolor{revision_1}{To extend TAU, we also remind the recommendation made in the introduction (Section 1): combining the use of UAVs and stationary cameras are the best approach to benefit from the advantages of both tools and overcome their limitations. For example, synchronizing TAU with a vehicle plate identification station installed on the road helps assign the real identity to every tracked vehicle. This opens the doors toward new possibilities like gathering information about the vehicle id, age, brand, and history. Then, better analysis and management of the traffic can be implemented.} \par
\textcolor{revision_1}{In the next version, TAU can also be extended by adding predictive analysis. An AI model can be developed to learn how to predict approximately the traffic status by learning from daily traffic data. This prediction greatly helps the traffic managers to avoid congestion and other traffic anomalies before happening. This helps more to discover the main factors that impact traffic fluidity for a better understanding of traffic. }\par
\textcolor{revision_1}{Apart from predictive analysis, TAU can be extended by enriching its ability to work on camera views. It is useful to investigate more the intrinsic and extrinsic parameters of the camera mounted in the UAV. This is to be able to deduce a strong foundation to extract traffic parameters from a camera view that is not perpendicular to the ground. Moreover, TAU should be assessed to work on the different weather conditions (day/night, sunny/rainy, etc.). This is primordial to ensure normal daily use.  } \par




\vspace{6pt} 








\section*{Acknowledgments}
The authors would like to acknowledge the support of Prince Sultan University for funding this study. Special acknowledgement to Robotics and Internet of Things Lab at Prince Sultan University, Riyadh, Saudi Arabia.

\bibliographystyle{./model2-names.bst}\biboptions{authoryear}
\bibliography{biblio}

\end{document}

\end{document}